\pdfoutput=1
\documentclass[11pt]{article}

\usepackage{acl}

\usepackage{times}
\usepackage{latexsym}
\usepackage{graphicx}
\usepackage{amsmath,amssymb,amsfonts}
\usepackage[T1]{fontenc}
\usepackage[utf8]{inputenc}

\usepackage{microtype}

\usepackage{inconsolata}

\title{Syntactic Structure Processing in the Brain while Listening}

\author{Subba Reddy Oota$^{1,2}$, Mounika Marreddy$^2$, Manish Gupta$^{2,3}$ and Bapi Raju Surampudi$^2$\\
$^1$INRIA, Bordeaux, France; $^2$IIIT Hyderabad, India; $^3$Microsoft, India\\
\texttt{\small subba-reddy.oota@inria.fr, mounika.marreddy@research.iiit.ac.in}\\
\texttt{\small gmanish@microsoft.com, raju.bapi@iiit.ac.in}
}

\begin{document}
\maketitle
\begin{abstract}
%Some recent neuroscience studies explore a new research direction to find brain areas for which the activity can be predicted by the syntactic representation of the stimulus words or sentences.

Syntactic parsing is the task of assigning a syntactic structure to a sentence. There are two popular syntactic parsing methods: constituency and dependency parsing. Recent works have used syntactic embeddings based on constituency trees, incremental top-down parsing, and other word syntactic features for brain activity prediction given the text stimuli to study how the syntax structure is represented in the brain's language network.
%To understand how syntax structure is represented in the brain's language network, recent works have used syntactic embeddings based on constituency trees, incremental top-down parsing, and other word syntactic features for brain activity prediction given text stimuli.
However, the effectiveness of dependency parse trees or the relative predictive power of the various syntax parsers across brain areas, especially for the listening task, is yet unexplored.
In this study, we investigate the predictive power of the brain encoding models in three settings: (i) individual performance of the constituency and dependency syntactic parsing based embedding methods, (ii) efficacy of these syntactic parsing based embedding methods when controlling for basic syntactic signals, (iii) relative effectiveness of each of the syntactic embedding methods when controlling for the other. Further, we explore the relative importance of syntactic information (from these syntactic embedding methods) versus semantic information using BERT embeddings. 
%Using the features of pretrained syntactic embeddings and story text (listening stories), 
%We model the brain representation of syntax for listening comprehension. 
We find that constituency parsers help explain activations in the temporal lobe and middle-frontal gyrus, while dependency parsers better encode syntactic structure in the angular gyrus and posterior cingulate cortex. Although semantic signals from BERT are more effective compared to any of the syntactic features or embedding methods, syntactic embedding methods explain additional variance for a few brain regions. 
% Our work leads to the following interesting findings. (i) Both constituency and dependency tree-based embeddings are effective across different language regions for brain activity prediction, even after controlling for basic syntactic signals. 
% %(ii) Incremental top-down parser based embeddings are ineffective.
% (ii) When compared in a pairwise manner, constituency parsers have superior performance for the activity in the temporal and middle-frontal gyrus, , while dependency parsers yield better accuracy in the angular gyrus and posterior cingulate cortex. (iii) Although semantic signals from BERT are more effective compared to any of the syntactic features or embedding methods, syntactic embedding methods explain additional variance for a few brain regions. 
% Finally, we also notice that brain activation for both syntax and semantic predictions overlap and are distributed across the language network for listening.
% graph convolutional networks (GCN)
% We also notice that brain activation for both syntax and semantic predictions overlap and are distributed across the language network for listening, unlike what is suggested by classical studies (for example, the syntax is specific to Brodmann's area BA 44).
% Vijay - redundant?
% The syntactic structure-based features from constituency parse trees are more relevant to how the brain processes syntax, implying a hierarchical structure.
% Finally, constituency parse trees have low semantic information than SynGCN, followed by incremental top-down parsers.
We make our code publicly available\footnote{\url{https://tinyurl.com/BrainSyntax}}.
%It is reasonable to assume that any additional variance predicted by the syntactic embeddings from three parsing methods compared to the semantic feature spaces BERT is mainly due to their syntactic information. 
%We also find that syntactic representations yield better predictive performance than semantic representations obtained from BERT.
%voxels are more different in left hemisphere compared to right hemisphere, are they are more distinct for a particular region?
%Prev sentence is incorrect. regions are same, voxels are different.
  %syntactic processing use effort-based metrics. 
%\textcolor{red}{Quantify how different. How much is the diff for reading or listening? Voxels are more different in left hemisphere. Among 4 regions, are they are more distinct for a particular region?}
\end{abstract}

\section{Introduction}
%The ability of word semantics and its syntactic representations enable the flexibility of human language.
%In particular, 
A key assumption in psycholinguistics is that sentence processing involves two operations: (i) the construction of a syntactic structure that represents the relation between its components and (ii) the retrieval of the meaning of single linguistic units from semantic memory. %(i.e., lexical items: words).
When presented with a sentence in a task, humans can understand word meaning effectively while reading and listening. Listeners and readers appear to extract a similar semantic meaning from narrative stories~\cite{rubin2000reading,diakidoy2005relationship}, hence suggesting that the brain represents semantic information in an amodal form, i.e., independent of input modality. Further, earlier language-fMRI encoding studies have observed that sentence semantics alone cannot explain all the variance in brain activity; \emph{syntactic} information can also be used to explain some of the variance~\cite{binder2016toward,fedorenko2014reworking}. 
%Which brain regions play a crucial role in processing both syntactic and semantic structure is a long-standing question in neuroscience. 
% \begin{figure*}[t]
% \centering
% \begin{minipage}{0.90\columnwidth}
% % \centering
% \hspace{20pt}
% \includegraphics[width=1.85\columnwidth]{images/arch.pdf}
% \end{minipage}
% \begin{minipage}[t]{0.90\columnwidth}
% \vspace{-100pt}
% \hspace{-0pt}
% \includegraphics[width=0.9\columnwidth]{images/trees1.pdf}
% \end{minipage}
% \caption{Four steps of our proposed approach: (1) fMRI acquisition, (2) Syntactic parsing, (3) Regression model training, and (4) Predictive power analysis of the three embeddings methods.
% \label{fig:arch}
% %Step 2 shows syntactic parsing example for the sentence ``The cat sat on the mat.'' (a) constituency parse tree, (b) dependency parse tree generated by a CoreNLP constituency parser~\cite{manning2014stanford}, and (c) incremental top-down parse tree.
% }
% \end{figure*}
\begin{figure*}[t]
\centering
\includegraphics[width=0.647\textwidth]{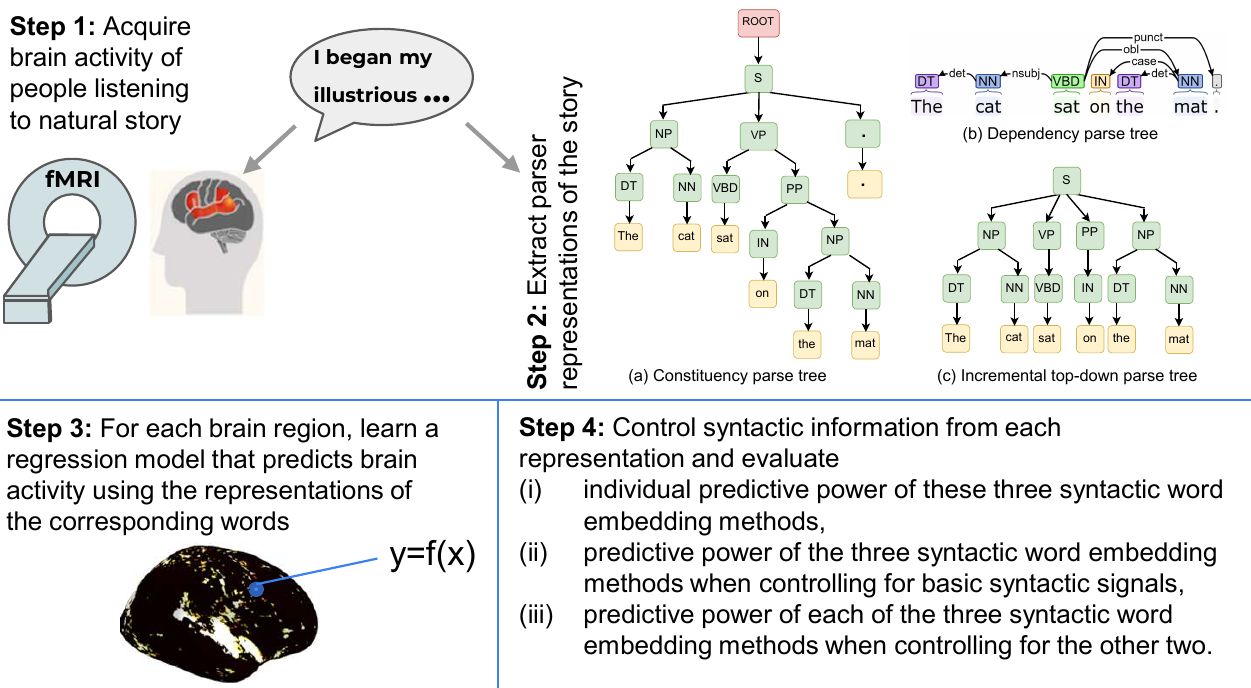}
\caption{Four steps of our proposed approach: (1) fMRI acquisition, (2) Syntactic parsing, (3) Regression model training, and (4) Predictive power analysis of the three embeddings methods.}
\label{fig:jointArch}
\end{figure*}

% \noindent\textbf{Importance of studying distinct brain encoding models of syntax: }
% As hypothesized in~\citet{rothstein1987syntactic}, the cognitive system assigns a unique syntactic structure to each word which is the input to semantic interpretation while processing strings of words.
Prior to different aspects of semantic interpretation,  the brain performs syntactic structure analysis inherently~\cite{hirst1984semantic}.
The syntactic information helps to identify the structural constituents that have to be interpreted as nominal, ordinal, or noun phrases, e.g., we identify ``Brazil'', ``four'', ``world cups'', and ``2002'' in a sentence: ``Brazil won four world cups till 2002'' before interpreting the semantics. Hence, investigating how the brain encodes syntactic word features is crucial for understanding language comprehension in the brain.

\noindent\textbf{Two paradigms of syntactic parsing:}
Constituency and dependency
%, and incremental top-down parsers
are two different syntactic formalisms using different structural primitives (dependency relations and phrases). There has been some discussion in the field of theoretical linguistics with regard to whether they capture the same information or to what degree the structures they sanction are equivalent~\cite{hays1964dependency,jung1995syntaktische}.
Discussing the linguistic information the two parsers capture,~\citet{rambow2010simple} states from a theoretical linguistic point of view that they describe distinct syntactic entities; thus, they are not strictly equivalent. Dependencies capture direct relations between words, identical to thematic functions such as subject, object, modifier, etc. Constituent syntactic structure, on the other hand, is not so much about functional relations between words but about the recursive grouping of sentence constituents (words and phrases), such that at each level, each grouping acts as a syntactic unit~\cite{schneider1998linguistic}. Moreover, according to~\citet{jung1995syntaktische}, only dependencies can express the syntactic word-to-word relations of a sentence, whereas constituency expresses the linear order of a sentence. 
On the other hand, an incremental top-down constituency parser processes input words from left to right, producing the possible parses in a top-down manner as future words are read.
Therefore, ~\citet{jung1995syntaktische} sees the two grammars as complementary but not equivalent. 
Following these last observations, we consider dependency and constituent structures as distinct and the type of information they capture as nonequivalent. 
% Although it is obvious from the above discussion that these parsers capture syntactic (and not semantic) information from sentences, we also verify it empirically in Section~\ref{sec:experiments}.
The question we address in this study is whether different brain regions are associated with building different kinds of syntactic structures. We compare the predictive power of syntactic structural measures derived from these parsers
%: dependency, constituency and incremental top-down parsers — 
with regard to modeling the brain activity in language processing areas recorded during naturalistic story listening.

\noindent\textbf{Stimulus types for studying syntactic processing:} 
Earlier psycholinguistic studies explored syntactic processing while subjects were involved in activities that required less versus more syntactic comprehension effort~\cite{friederici2011brain} using carefully designed sentence/phrase stimuli.
In the past decade, the study of syntactic processing has been extended to naturalistic settings that use narratives, such as reading ~\cite{reddy2021can} or listening to stories ~\cite{bhattasali2018differentiating,
%lopopolo2021distinguishing,
zhang2022probing} generally in a task-free setting.
%~\cite{brennan2012syntactic,willems2016prediction}. 
%In particular, most of the recent studies have attempted to address the syntactic structure during reading stories~\cite{reddy2021can}, 
Due to the complexity of extracting syntactic word embeddings from sentence parsers, investigation of the predictive power of sentence parsers for brain encoding, especially for the neuroimaging data from naturalistic listening paradigms, is still under-explored.

%However, the problem of understanding the distribution of syntactic processing across brain regions when listening to stories is still under-explored, and is therefore the focus of this paper.
%Another unanswered question is whether the brain processes syntax information similarly while reading and listening. 

%However, to date, few studies have attempted to shown that regions in the left-hemisphere were highly involved in semantics of language processing~\cite{vigneau2006meta,binder2009semantic}. 
\noindent\textbf{Brain Regions of Interest (ROIs) for syntactic processing:}
%Since the discovery of the relationship between language stimuli and functions of brain networks using fMRI~\citep{constable2004sentence}, researchers have been interested in identifying brain ROIs that respond to words with similar meanings and have thus built a ``semantic atlas'' of how the human brain organizes language~\cite{huth2016natural}. 
%Several classical studies used fMRI data to show that few brain regions (BA 44 and BA 45) are specific to syntax processing~\cite{friederici2011brain}.
Several classical studies report the involvement of a language network of mostly left-lateralised cortical regions, including the left inferior frontal gyrus (IFG) with sub-regions (BA 44 and BA 45), the left posterior superior temporal gyrus (pSTG), and the left anterior temporal pole (ATP)~\cite{caramazza1976dissociation,friederici2006processing,friederici2011brain,pallier2011cortical,zaccarella2015merge}. 
However, several other studies did not report activity in left IFG and left pSTG~\cite{humphries2006syntactic,rogalsky2009selective,bemis2011simple}, despite using paradigms similar to the studies mentioned above.
A series of recent studies have used functional magnetic resonance imaging (fMRI) brain activity to find that those brain regions spanning both the left and right hemispheres are involved in language processing~\cite{fedorenko2014reworking,caucheteux2021disentangling,reddy2021can,zhang2022probing,oota2022neural,oota2022visio,oota2022long,toneva2022combining,aw2022training,oota2022joint,merlin2022language}. 
Further, these works conclude that syntax is distributed throughout the language system~\cite{blank2016syntactic,fedorenko2012lexical,fedorenko2020lack,caucheteux2021disentangling,wang2020probing,reddy2021can,zhang2022probing,oota2022neural}.
However, whether different brain regions are sensitive to distinct sentence-parsing strategies remains unclear. Moreover, in a listening task, it is unclear how syntactic features are represented in the brain and whether the neural correlates of different syntactic parsing signals overlap or dissociate from one another.

%However, which brain areas are essential for syntactic processing \emph{while listening} is unknown. We attempt to fill the gap in this paper.
%\textcolor{red}{Cite this paper: wang2020probing}

%Recent advances in brain imaging studies suggest that it may be possible to investigate how syntax structure is represented in Brain~\cite{reddy2021can}.

%and which parsing method is more relevant to brain.

%Why do we need sentence structure? 
%In particular, the syntactic parsing is task of identifying structure ambiguity to a sentence. In the linguistic studies, there are three parsers which mainly used for parsing linguistic structure: (i) constituency parsing, (ii) dependency parsing, and (iii) incremental top-down parsing. Among three parsers, both constituency tree and incremental top-down tree provide full syntactic analysis, minimally a constituency (phrase-structure tree) parse of sentences.
%Inorder to get syntactic structure using dependency trees, popular graph neural networks have been used to process syntactic
%information. With the popular tree parsers, our main objective is to understand how human brain process sentences to interpret the linguistic structure correctly and which parsing method is more relevant to brain.

\noindent\textbf{Word stimulus representations for brain encoding}: 
Several studies have used basic syntactic features such as part-of-speech, dependency relations, complexity metrics~\cite{caucheteux2021disentangling,reddy2021can}, and semantic word embeddings~\cite{oota2018fmri,jain2018incorporating,hollenstein2019cognival} to represent words for brain encoding with text stimulus.
%where words are mapped to brain activation for estimation of neural encoding. 
% Recently,~\citet{zhang2022probing} explored the syntactic representations by feature elimination method that can remove a specific feature from word embeddings while retaining other features. 
% In the field of Natural Language Processing (NLP), parsing sentences into syntax trees has benefited several applications such as text generation~\cite{kumar2020syntax}, summarization~\cite{li2014improving,song2020joint}, and machine translation~\cite{gu2018top}. 
In this paper, to understand how the brain processes linguistic structure in sentences, we leverage three different text representations using syntax parsers, as shown in Fig.~\ref{fig:jointArch}. We aim to understand the relative importance of these syntax parser embeddings and also their additional importance when compared with basic syntactic features or semantic embeddings like BERT.
%(i) constituency parsing~\cite{gaddy2018s}, (ii) dependency parsing~\cite{covington2001fundamental}, and (iii) incremental top-down parsing~\cite{roark2001probabilistic} 
% We empirically show in Section~\ref{sec:experiments} that these representations are indeed syntactic and do not encode much semantics.
% Among the three parsers, constituency and incremental top-down parsing provide a complete syntactic analysis. 
% In order to effectively extract syntactic structure using dependency parsing, graph neural networks have been popularly used.
% We used random walk with paragraph vectors (sub2vec) approach suggested in~\cite{} for constituency and incremental top-down parsing.

\noindent\textbf{Limitations of previous work}:  
(i) Existing work has focused on either constituency parsing mainly including incremental top-down parsing~\cite{reddy2021can}. No previous work has explored syntactic structure present in dependency trees.~\citet{reddy2021can} have only used one-hot vector for dependency tags as part of their complexity metrics. But we leverage dependency information more systematically by learning the dependency representations using graph convolutional networks.
% using a graph embeddings-based approach.
% (iii) Investigating the importance of elementary syntactic features by removing a specific feature from the original word representations is not generalizable since every feature elimination results in a different syntactic embedding~\cite{zhang2022probing}. 
(ii) Existing work has mostly focused on reading tasks only, and that too on small number of subjects (e.g., 7 subjects in~\cite{reddy2021can}). There is evidence that several cortical regions are activated during listening~\cite{handjaras2016concepts}. But which brain areas and sub-regions of the language network are involved in syntactic processing is yet unexplored. (iii) Lastly, existing work does not perform pairwise predictive power comparison for different syntactic parse methods. %Further, it is not yet clear which different sub-regions of the language network are engaged in different aspects of syntax processing. 
% and reading. 
% However, demonstrating some common activation is necessary but not sufficient evidence of a common amodal semantic representation. 

%Like semantic representation, does the regions of brain perform similar or different syntactic representations 

Overall, our main contributions are as follows. 
% \begin{enumerate}
    (1) We explore (a) basic syntactic features such as complexity metrics, part-of-speech (POS) tags, and dependency role (DT) tags, (b) embeddings obtained from three parse tree representations, and (c) semantic BERT embeddings for brain encoding. 
    %We use a GCN model (DEP embeddings) for the dependency parser that accurately encodes the global syntactic information.
    %We use a random-walk with paragraph vectors (sub2vec) approach for the constituency and top-down parsers to generate the embeddings~\cite{reddy2021can}. 
    (2) %In Section~\ref{removal properties}, we show that the removal of each linguistic property in the BERT representations results in reduced probing class performance across all the layers. Also, 
    %Investigation of individual predictive power of representations shows that 
    Constituency and dependency tree-based embeddings are effective across different language regions for brain activity prediction, even after controlling for basic syntactic signals.
    (3) %We conduct a detailed study of alignment of the representations with specific brain regions that are thought to underlie language comprehension. 
    We find that prediction of the activation in regions such as the bilateral temporal areas (ATL, PTL) and middle-frontal gyrus (MFG) is significantly related to constituency parse representations. At the same time, brain activity in other language regions, such as the angular gyrus (AG) and posterior cingulate cortex (PCC) is significantly associated with dependency parse embeddings.
    (4) Lastly, in the inferior frontal gyrus (IFG), we identify that dependency parse embeddings encode syntactic information better in the sub-regions such as 44, 45, IFJa, and IFSp of the left hemisphere, whereas constituency parse tree and incremental top-down parse tree based embeddings are better aligned in the right hemisphere.
% \end{enumerate}

%\noindent\textbf{Our main contributions: }
%better using graph convolutional networks (GCNs) and do not restrict graphs to be trees and have been found to be more effective at capturing global information.
% These syntactic embeddings can serve as an additional tool in the arsenal of Neurolinguists to explore whether the fMRI signal can reveal syntactic representations.
%In particular,   We use both pretrained SynGCN embeddings trained on Wikipedia as well as we generated from scratch on reading and narrative stories. 

% We study the syntactic structure in the brain while subjects are listening to narrative stories and thus provide insights around brain ROIs involved in syntactic processing for listening tasks.

\vspace{-0.1cm}
\section{Feature Representations}
We used four different features computed per word to simultaneously test different syntactic and semantic representations.

\noindent\textbf{(1) Constituency Tree-based Embeddings: }
Similar to~\citet{reddy2021can}, we build three types of constituency tree-based graph embeddings (ConTreGE): (i) ConTreGE Complete vectors (CC), (ii) ConTreGE Incomplete vectors (CI) and (iii) Incremental Top-Down Parser Embeddings (INC). 
A CC vector is generated for every word using the largest subtree completed by that word. 
A subtree is considered complete when all of its leaves are terminals. 
The largest subtree completed by a given word refers to the subtree with the largest height. 
% that also satisfies the following conditions:
% 1. The given word must be one of its leaves,
% 2. All of its leaves must only contain words that have been seen till then.
A CI vector is generated for every word using the incomplete subtree that contains all of the Phrase Structure Grammar productions needed to derive the words seen till then, starting from the root of the sentence’s tree. Some examples for CC and CI are added in the Appendix (Figs.~\ref{fig:complete_tree} and~\ref{fig:incomplete_tree}). Like~\cite{reddy2021can}, we use Berkeley Neural Parser\footnote{\url{https://spacy.io/universe/project/self-attentive-parser}\label{note1}} for constituency parsing (i.e., for both CI and CC).

In ConTreGE Complete tree (CC), the largest subtree completed by a given word refers to the subtree with the largest height that also satisfies the following conditions - the given word must be one of its leaves and all of its leaves must only contain words that have been seen till then.

In ConTreGE Incomplete tree (CI), the embeddings are constructed using incomplete subtrees that are constructed by retaining all the phrase structure grammar productions that are required to derive the words seen till then, starting from the root of the sentence’s tree. If incomplete subtrees are more representative of the brain’s processes, it would mean that the brain correctly predicts certain phrase structures even before the entire phrase or sentence is read.

% \paragraph{SynGCN (WikiPedia)}
% Here, we use the pretrained SynGCN word embeddings trained on Wikipedia data~\cite{vashishth2019incorporating}.
% We prefer the word embeddings from GCNs over graph embeddings which use parse trees because SynGCN method encode syntactic information better using GCNs and do not restrict graphs to be trees and have been found to be more effective at capturing global information~\cite{zhang2018graph}.
% Moreover, SynGCNs employ edge-wise gating mechanism~\cite{marcheggiani2017encoding} that provides a relevance score for all the edges.
% The representation of each word obtained using pretrained SynGCN embeddings where each word denotes a 300 dimensional vector.
The incremental top-down parser is a statistical syntactic parser that
processes input strings from left to right, producing partial
derivations in a top-down manner, using beam search as detailed in~\cite{roark2001probabilistic}.  Specifically, we use the implementation as described here\footnote{\url{https://github.com/roarkbr/incremental-top-down-parser}\label{note2}}. The INC embeddings are obtained using exactly the same methods as described in Section 3 of~\cite{reddy2021can}. 
% It can output parser state statistics of utility for psycholinguistic studies, as detailed in Roark et al. (2009).
The brain could be computing several possible top-down partial parses that can derive the words seen so far and modifying the list of possible parses as future words are read.
The INC feature space is constructed to encode the different possible parse trees that can derive the words seen so far. When considering parse tree based representations, the embeddings may contain information about what is yet to be seen by the subject. However, this is not a problem since it mimics the human capability of guessing what is to come next. With this embedding space, we attempt to measure the ability of the brain to predict future constituents correctly. 

\noindent\textbf{(2) Dependency Tree-based Embeddings (DEP): }
Graph Convolutional Networks (GCNs) have been widely used to encode syntactic information from dependency parse trees~\cite{vashishth2019incorporating}.
%in NLP to process syntactic information. These methods derive syntax embeddings by applying GCNs on dependency parse trees~\cite{vashishth2019incorporating}.
% for downstream tasks such as pronoun resolution
Rather than using pretrained syntactic GCN word embeddings generated from Wikipedia~\cite{vashishth2019incorporating}, we create DEP embeddings using GCNs on the ``Narrative stories'' dataset as follows. 
To generate syntactic word embeddings using GCN, 
we first extract the dependency parse tree $G_s$=($V_s$, $\epsilon_s$) 
for every sentence in our dataset $s$ = ($w_1, w_2,$\dots$, w_n$), 
using the Stanford CoreNLP parser~\cite{manning2014stanford}. 
Here, $V_s$ = \{$w_1$, $w_2$,$\dots$, $w_n$\} and $\epsilon_s$ denotes the labeled directed dependency edges of the form ($w_i$, $w_j$ , $l_{ij}$ ), where $l_{ij}$ is the dependency relation of $w_i$ to $w_j$.
% Now, unlike the CBOW word embedding method which takes the sum of the context embedding of words in a small window to predict embedding for word $w_i$, 
GCN computations iteratively utilize the context defined by a word's neighbors in the graph to compute embedding for every word $w_i$.
Further, we also perform edge-wise gating to give importance to relevant edges and suppress noisy ones. We follow the architecture defined in~\citep{vashishth2019incorporating} for training a GCN on our dataset leading to syntactically-rich DEP embeddings.  
Overall, GCN utilizes syntactic context to learn rich DEP embeddings.

% \paragraph{SemGCN}
% For incorporating semantic knowledge into pretrained SynGCN embeddings, semantic relations such as hyponymy, hypernymy and synonymy are are introduced between different nodes i.e edges representing semantic relation- ship among them from different sources.
% \begin{figure}[t] 
% \centering
% \includegraphics[width=0.9\linewidth]{images/SynGCN.pdf}
% \caption{SynGCN embeddings}
% \label{fig:syngcn}
% \end{figure}

\noindent\textbf{(3) Basic Syntactic Features:}
Similar to~\citep{wang2020probing,reddy2021can,zhang2022probing}, we use various multi-dimensional syntactic features such as Punctuation (PU), Complexity Metrics (CM), and Part-of-speech and dependency tags (PD), described briefly below.

\noindent\textbf{Punctuation (PU)}
The role of punctuation is to resolve syntactic and semantic ambiguity in the lexical grammar and encode relational discourse links between text units in sentences~\cite{briscoe1996syntax}. 
% For example, `tone' indicators such as question (?) and exclamation marks (!) serve to alter or emphasise grammatical mood and information structure. 
Punctuation-based features are encoded using a one-hot vector where the type of punctuation is presented along with a word (e.g. . or ,).

\noindent\textbf{Complexity Metrics (CM)}
We use three features in the complexity metrics: Node Count (NC), Word Length (WL), and Word Frequency (WF). 
The node count for each word is the number of subtrees that are completed by incorporating each word into its sentence.
Word length is the number of characters present in the word.
%We use the Zipf frequency to measure the frequency of each word. 
Word frequency reports log base-10 of the number of occurrences per billion of a given word in a large text corpus.

\noindent\textbf{Part-of-speech and Dependency tags (PD)}
We use the Spacy English dependency parser~\cite{spacy2} to extract the Part-of-speech (POS) and dependency tags. 
Unlike DEP embeddings (which use GCNs), in PD, we generate a one-hot vector for each word and dependency tag.
%in which corresponding POS tag location is 1 and remaining tag values are 0. 
%Similarly, we create one-hot vector for dependency tags. 
The final vector is called PD, a concatenation of both the POS tag and dependency vector. Note that DEP and PD features use different methods for dependency analysis -- PD features are just one-hot encoded representations while DEP features are learned syntactic embeddings using GCNs.

\noindent\textbf{(4) BERT Features}
Given an input sentence, the pretrained  BERT~\cite{devlin2018bert} outputs token representations at each layer. 
Since BERT embeds a rich hierarchy of linguistic signals: surface information at the bottom, syntactic information in the middle, semantic information at the top~\cite{jawahar2019does}; hence, we use the \#tokens $\times$ 768D vector from the last hidden layer to obtain the semantic embeddings.
%We use the \#tokens $\times$ 768 dimension vector from the last hidden layer as latent features for the stimuli. 
For uniformity of feature dimensions, we used PCA to bring down the dimensions to 250.

%\vspace{-0.1cm}
\section{Dataset Curation}
\vspace{-0.1cm}

\noindent\textbf{Brain Imaging Dataset}
\label{sec:dataset}
%We work with the Narratives-Pieman~\cite{nastase2021narratives} story listening dataset.
% \noindent\textbf{\emph{Harry Potter} Dataset (Reading Story)} 
% The \emph{Harry Potter} fMRI dataset collected while human subjects were reading a story of Harry Potter and the Sorcerer’s Stone~\citet{brown2002harry}. The Harry Potter dataset includes 9 subjects, words are presented sequentially for 0.5s each.
% We briefly summarize the details of the dataset and the number of voxels corresponding to each subject in Table~\ref{tab:harrypotter_stats} (refer to appendix). We use the AAL parcellation Atlas (116 $\times$ 116 brain ROIs) to present the brain map results, since Harry Potter dataset contains annotations tied to this atlas. The dataset made available freely without restrictions by ~\citet{wehbe2014simultaneously}.
%\noindent\textbf{Narratives-Pieman (Listening to Stories)} 
The ``Narratives'' collection aggregates a variety of fMRI datasets collected while human subjects listened to real spoken stories~\cite{nastase2021narratives}. 
%The Narratives dataset includes 345 subjects, 891 functional scans, and 27 diverse stories of varying duration, totaling $\sim$4.6 hours of unique stimuli ($\sim$43,000 words). Similar to earlier works~\cite{caucheteux2021model}, 
We analyze data from 82 subjects listening to the story titled `PieMan' with 282 TRs (repetition time -- fMRI recorded every 1.5 sec.). We chose this story since it contains maximum number of subjects in the ``Narratives'' collection. The dataset is in English and contains 957 words across 67 sentences. The story duration is 07m:02s. We use the multi-modal parcellation of the human cerebral cortex (Glasser Atlas: consists of 180 ROIs in each hemisphere) to display the brain maps~\cite{glasser2016multi} since the Narratives dataset contains annotations tied to this atlas. 
The data covers eight language brain ROIs with the following subdivisions: (i) angular gyrus (AG: PFm, PGs, PGi, TPOJ2, and TPOJ3); (ii) anterior temporal lobe (ATL: STSda, STSva, STGa, TE1a, TE2a, TGv, and TGd); (iii) posterior temporal lobe (PTL:   A5, STSdp, STSvp, PSL, STV, TPOJ1); (iv) inferior frontal gyrus (IFG: 44, 45, IFJa, IFSp); (v) middle frontal gyrus (MFG: 55b); (vi) inferior frontal gyrus orbital (IFGOrb: a47r, p47r, a9-46v), (vii) posterior cingulate cortex (PCC: 31pv, 31pd, PCV, 7m, 23, RSC); and (viii) dorsal medial prefrontal cortex (dmPFC: 9m, 10d, d32)~\cite{baker2018connectomic,milton2021parcellation,desai2022proper}.
%We report the mapping of ROIs corresponding to the language network in the appendix (please see Table~\ref{tab:roismapping}). 
The dataset has been made available freely without restrictions by~\citet{nastase2021narratives}.

\noindent\textbf{Downsampling}
Since the rate of fMRI data acquisition (TR = 1.5sec) was lower than the rate at which the text stimulus was presented to the subjects, several words fall under the same TR in a single acquisition. 
Hence, we match the stimulus acquisition rate to fMRI data recording by downsampling the stimulus features using a 3-lobed Lanczos filter~\cite{lebel2021voxelwise}.
After downsampling, we average the word-embeddings within each TR to obtain chunk-embedding for each TR. 

\noindent\textbf{TR Alignment}
To account for the slowness of the hemodynamic response, we model the hemodynamic response function using finite response filter (FIR) per voxel and for each subject separately with 8 temporal delays corresponding to 12 seconds. 

%\vspace{-0.1cm}
\section{Methodology}
\label{headings}
\vspace{-0.1cm}
\noindent\textbf{Encoding Model} 
To explore how and where syntactic and semantic specific features are represented in the brain when listening to stories, we extract different features describing each stimulus sentence and use them in an encoding model to predict brain responses.
If a feature is a good predictor of a specific brain region, information about that feature is likely encoded in that region.
% In particular, the applicability of a given syntactic feature in studying syntactic processing is determined by its efficacy in predicting brain activity.

The main goal of each fMRI encoder model is to predict brain responses associated with each brain voxel when given stimuli. 
We train a model per subject separately. Following the literature on brain encoding~\cite{wehbe2014simultaneously,toneva2020modeling,caucheteux2021model, reddy2021can,toneva2021same,zhang2022probing,oota2022neural,oota2022visio}, we choose to use a ridge regression model instead of more complicated models. 
We plan to explore more such models as part of future work. %In this paper, we train fMRI encoding models using ridge regression on stimuli representations.
The ridge regression objective function for the $i^{th}$ example is $f(X_{i})= \underset{W}{\text{min}} \lVert Y_i - X_{i}W \rVert_{F}^{2} + \lambda \lVert W \rVert_{F}^{2}$.
% \begin{align*}
% \end{align*}
Here, $W$ are the learnable weight parameters, $\lVert.\rVert_{F}$ denotes the Frobenius norm, and $\lambda >0$ is a tunable hyper-parameter representing the regularization weight. $\lambda$ was tuned on a small disjoint validation set obtained from the training set.

\noindent\textbf{Cross-Validation}
We follow 4-fold (K=4) cross-validation. All the data samples from K-1 folds were used for training, and the model was tested on samples of the left-out fold. 

\noindent\textbf{Evaluation Metric} We evaluate our models using the popular brain encoding evaluation metric, $R^2$. Let TR be the number of time repetitions. Let $Y=\{Y_i\}_{i=1}^{TR}$ and $\hat{Y}=\{\hat{Y}_i\}_{i=1}^{TR}$ denote the actual and predicted value vectors for a single voxel. Thus, $Y\in R^{TR}$ and also $\hat{Y}\in R^{TR}$. 
We use R$^{2}(Y, \hat{Y})$ metric to measure the coefficient of determination for every voxel. 
%We also use Pearson Correlation (PC) which is computed as $corr(Y, \hat{Y})$ where corr is the correlation function. 
We average $R^2$ score over all voxels in a region to get region-level aggregated metric. Finally, they are further averaged across all subjects to obtain final region-level metrics, which are reported with mean and standard deviation. % in Figs.~\ref{fig:listening_syngcn_all_left} (\% of significant voxels) and~\ref{fig:listening_avg_r2_allmodels} (average $R^2$ score).
% and~\ref{fig:lefthemisphere_corr}.
%\textcolor{red}{We never report corr coeff then why mention it here.}

\noindent\textbf{Statistical Significance}
We run a permutation test to check if $R^2$ scores are significantly higher than chance. We permute blocks of contiguous fMRI TRs, instead of individual TRs, to account for the slowness of the underlying hemodynamic response. 
We choose a standard value of 10 TRs. 
The predictions are permuted within fold 5000 times, and the resulting $R^2$ scores are used as an empirical distribution of chance performance, from which the p-value of the unpermuted performance is estimated. 
We also run a bootstrap test, to test if a model has a higher $R^2$ score than another. 
In each iteration, we sample with replacement the predictions of both models for a block of TRs, compute the difference of their $R^2$, and use the resulting distribution to estimate the p-value of the unpermuted difference. 
Finally, the Benjamni-Hochberg False Discovery Rate (FDR) correction~\cite{benjamini1995controlling} is used for all tests (appropriate because fMRI data is considered to have positive dependence~\cite{genovese2000bayesian}). 
The correction is performed by grouping all the voxel-level p-values (i.e., across all subjects and feature groups) and choosing one threshold for all of our results. 
The correction is done this way as we test multiple prediction models across multiple voxels and subjects.

\begin{figure*}[!t] 
\centering
%\vspace{-1cm}
\includegraphics[width=0.65\linewidth]{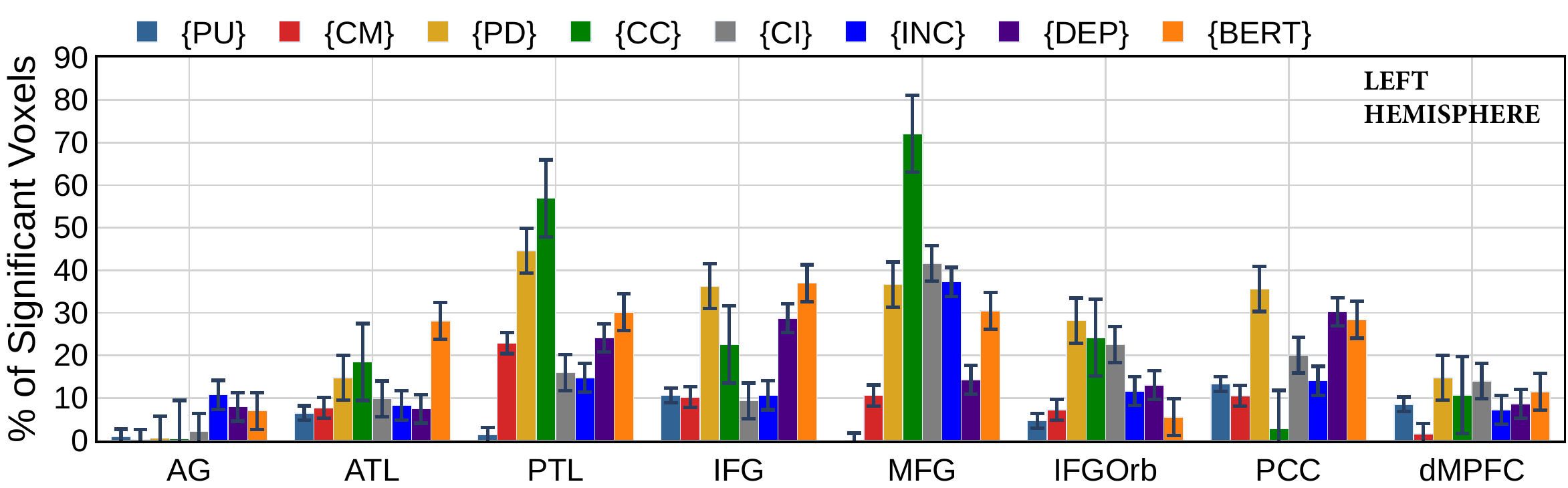}
%\vspace{-0.3cm}
\includegraphics[width=0.65\linewidth]{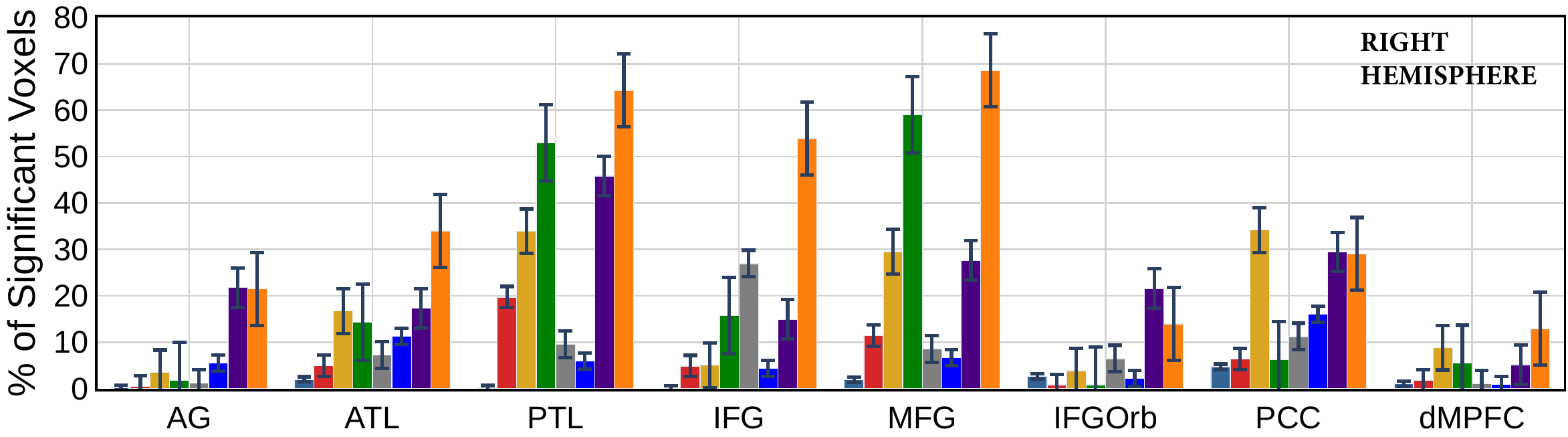}
\vspace{-0.3cm}
\caption{\textbf{Performance of Individual Embedding Methods}: ROI-wise analysis of the prediction performance of various feature sets. We show the \% of ROI voxels with a significant increase in prediction performance. Each bar shows avg \%; error bars show standard error across 82 subjects. Left hemisphere (Top); Right hemisphere (Bottom).}
%‘-’ indicates a hypothesis test for the difference in $R^2$ scores between the two feature groups being larger than 0.
\label{fig:listening_syngcn_all_left}
\end{figure*}

% We compute these metrics per subject and report the average and std deviation in our results. 

% \textcolor{red}{Manish to handle.}
% Given and a brain region,  Thus, $Y\in R^{TR\times V}$ and $\hat{Y}\in R^{TR\times V}$ where $V$ is the number of voxels in that region. We use R$^{2}(Y, \hat{Y})$ metric\footnote{\url{https://scikit-learn.org/stable/modules/generated/sklearn.metrics.r2\_score.html}} to measure the coefficient of determination for every voxel. We also evaluate Pearson Correlation (PC) which is computed as PC=$\frac{1}{N}\sum_{i=1}^{n} corr[Y_i, \hat{Y}_i]$ where corr is the correlation function.

%\noindent\textbf{R$^{2}$} is computed for every voxel as R$^{2}$=1-$\frac{1}{N}\sum_{i=1}^{n} |[Y_i-\hat{Y}_i]|$.

% \begin{figure*}[!htb] 
% \centering
% \includegraphics[width=\linewidth]{images/narratives_syngcn_all_righthemisphere.pdf}
% \caption{Right-Hemisphere Region of Interest (ROI) analysis of the prediction performance of various feature sets. For each model, we show the percentage of ROI voxels in which we see significant increase in prediction performance. Each bar represents the average percentage across 82 subjects and the error bars show the standard error across subjects. ‘-’ indicates a hypothesis test for the difference in $R^2$ scores between the two feature groups being larger than 0.}
% \label{fig:listening_syngcn_all_right}
% \end{figure*}

\begin{figure*}
\centering
\includegraphics[width=0.7\linewidth]{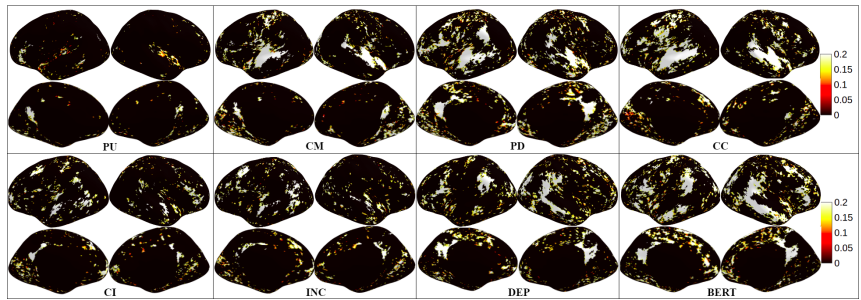}
\caption{$R^2$ score per voxel for the whole brain.
%comprising the Language Network as per~\citet{glasser2016multi}'s atlas. 
(a) PU (b) CM (c) PD (d) CC (e) CI (f) INC (g) DEP (h) BERT.}
\label{fig:listening_left}
\end{figure*}

\section{Experiments and Results}
\label{sec:experiments}
We discuss detailed hyper-parameter settings in  Appendix~\ref{sec:hyperParameters}.
%\subsection{Results}
%In this section, we attempt to answer the following questions:

% \begin{itemize}
%     \setlength\itemsep{-0.5em}
%     \item Is the representation of syntactic structure in the brain similar for different syntactic representations?
%     \item Which syntactic parsing method encodes syntactic structure information better aligned with fMRI brain activity?
%     \item Are syntax and semantics processed in a distributed way in language system for listening?
% \end{itemize}

\noindent\textbf{Which word representations are semantic versus syntactic?} We first empirically show that syntactic embeddings do not encode a significant amount of semantic information.
In particular, we train the RidgeCV regression model in a 10-fold cross-validation setting to predict the semantic GloVe~\cite{pennington2014glove} features (300 dimensions) using syntactic embeddings for all the representations, similar to earlier works~\cite{caucheteux2021disentangling,reddy2021can,zhang2022probing}.

Average $R^2$ scores are as follows: BERT (0.560), CC (0.052), CI (0.020), DEP (0.170), INC (0.040), PD (0.183), CM (0.027), and PU (0.005). These $R^2$ scores indicate that (a) overall, BERT has high semantic information compared to other embeddings, and (b) constituency parsers have low semantic information compared to DEP.
Overall, all the syntactic embeddings consist of very low semantic information. Hence, it is reasonable to infer that any additional variance predicted by the syntactic parsing methods compared to the semantic feature space (BERT) is mainly due to their syntactic information. 

\noindent\textbf{Performance of individual embedding methods:}
In order to assess the performance of the fMRI encoder models learned using the individual syntactic and semantic representations, we computed the $R^2$ scores between the predicted and true responses across various ROIs of the language network.
Fig.~\ref{fig:listening_syngcn_all_left} reports the \% of ROI voxels with significant $R^2$ scores (based on a hypothesis test where the $R^2$ score for each voxel is greater than 0) across different representations for different language regions in the left and right hemispheres. 
We make the following observations from Fig.~\ref{fig:listening_syngcn_all_left}: (1) Among basic syntactic features, PD features perform best across most of the language regions, whereas CM yields the second-best result. (2) Among the syntactic embedding methods, CC encodes syntactic information better in the language regions such as temporal lobes (ATL and PTL) and MFG. (3) Among the syntactic embedding methods, DEP embeddings predict brain activity better in the language regions (PCC and IFG of left hemisphere, and AG, IFGorb, and PCC of right hemisphere). (4) Semantic embeddings using BERT are the best across all regions in the right hemisphere, but the effectiveness of BERT is rather mixed in the left hemisphere.

Further, we report the avg $R^2$ scores across all different language ROIs in the Appendix (Fig.~\ref{fig:listening_avg_r2_allmodels}). 
We further demonstrate the performance of embedding methods for various sub-regions of each language ROI in the Appendix  Figs.~\ref{fig:listening_avg_r2_AG} to~\ref{fig:listening_avg_r2_dmPFC}. We observe the following from these figures: (1) In the ATL region (Fig.~\ref{fig:listening_avg_r2_ATL}), CC better encodes in the superior temporal sulcus with dorsal attention (STSda). For STS in ventral attention (STSva), CC encodes better in the left hemisphere while DEP is better in the right. (2) In the PTL region (Fig.~\ref{fig:listening_avg_r2_PTL}), CC is best for STSdp sub-region. (3) In the IFG region (Fig.~\ref{fig:listening_avg_r2_ifg}), DEP is better aligned with 44 region whereas CC is better aligned with IFJa region. These results are in line with observations made in~\cite{pallier2011cortical}.
% These results add that, in STSda, STSdp, and IFJa where fMRI responses get increasingly larger for constituent parsers~\cite{pallier2011cortical} while STSva, STSvp, 44, and 45 are for DEP parser.
%\textcolor{red}{compared to what? Also we should write about right hemisphere also?}
Overall, a higher percentage of voxels with all the frontal and temporal regions, demonstrates that language comprehension may be associated more with both frontal and temporal regions~\cite{cohen2021does}.
%\textcolor{red}{PTL also seems to have high correlation values, no?}
\begin{figure*}[!t] 
\centering
%\vspace{-1cm}
\includegraphics[width=0.78\linewidth]{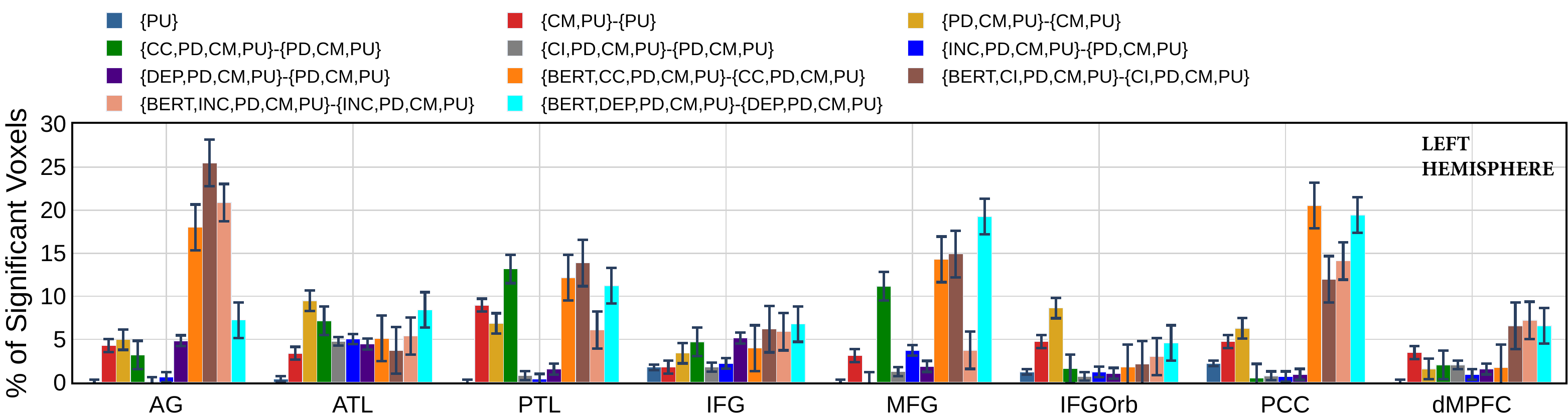}
%\vspace{-0.3cm}
\includegraphics[width=0.78\linewidth]{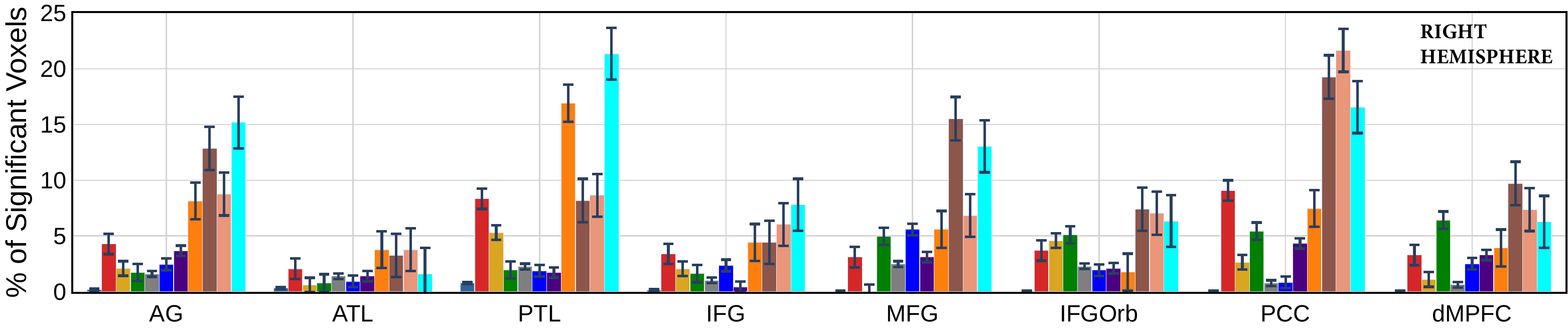}
%\vspace{-1cm}
% Hierarchical feature groups
% Controlling for syntactic information.
\caption{\textbf{Additional Predictive Power of various Representations}: For each model, we show the \% of ROI voxels with a significant increase in prediction performance. Each bar shows avg \%; error bars show standard error across 82 subjects. ‘-’ indicates a hypothesis test for the difference in $R^2$ scores between the two feature groups being larger than 0. Left hemisphere (Top); Right hemisphere (Bottom). Note that PU values here are slightly different from Fig.~\ref{fig:listening_syngcn_all_left} since here the FDR correction was done across all the groups.}
\label{fig:listening_syngcn_all_left_pairs}
\end{figure*}

% We make the following observations from Fig.~\ref{fig:lefthemisphere_corr}: (1) INC has higher correlation score compared to other methods. (2) PU and CM features display lower correlation compared to the remaining methods. 
%Fig.~\ref{fig:brainmaps_cc_ci_inc_syngcn} shows the mean Pearson correlation between actual and predicted voxels for various syntactic representations. 
We also report brain maps with avg $R^2$ for all the representations in  Fig.~\ref{fig:listening_left}. 
%\textcolor{red}{We must swap positions of Figures 2 and 3 for sure.} \textcolor{red}{what is plotted in these brain maps?}
From Figs.~\ref{fig:listening_syngcn_all_left} and~\ref{fig:listening_left}, we can infer that the different word representations, including all syntactic and semantic methods, are highly distributed across ROIs of language network. In particular, PTL and MFG have high overlap for both syntactic (CC, CI, DEP, INC), and semantic (BERT) features. Also, ROIs such as PTL, IFGOrb and PCC have higher overlap with PD. Most of these observations agree with previous findings on the brain networks of language processing~\cite{friederici2011brain,fedorenko2014reworking,caucheteux2021disentangling,reddy2021can,zhang2022probing}, supporting that both syntax and semantics are distributed across language ROIs. Lastly, similar to an earlier study~\cite{blank2016syntactic}, basic syntactic features are much less associated with voxels in AG region.
%validity of our detailed study that uses different syntactic parsing and semantic representations to study language comprehension in the brain. \textcolor{red}{not sure if this desc seems interesting. Either we need to write something interesting here or move this to appendix.}

\noindent\textbf{Additional predictive power of various representations}
%To investigate whether the syntactic structure represented in the brain is similar for different feature representations, we compare the voxels with $R^2$ scores across different language regions for narrative listening. 
%Fig~\ref{fig:listening_wholebrain} displays language regions for left and right hemispheres. 
Many feature spaces have overlapping information, e.g., PD (part-of-speech and dependency) tags include punctuation, BERT vectors have been shown to encode syntax~\cite{jawahar2019does,luoma2020exploring}, and DEP embeddings built from GCNs encode some POS tags information. Are various representations capturing very similar signals, i.e., redundant or capturing new information, which is additionally useful to predict brain activations? To answer this question, we first organize the feature groups in the increasing order of syntactic information. 
%We use different sets of feature spaces as inputs for encoding models that predict activity in each fMRI voxel. 
% To detect brain regions sensitive to the specific information given by a feature space, 
We build hierarchical feature groups in increasing order of syntactic information and test for significant differences in prediction performance between two consecutive groups. We start with the simple feature -- punctuation (PU) and then add more complex features in this order: the complexity metrics (CM), POS and dependency tags (PD), \{CC, CI, INC, DEP\}, and lastly, BERT. Fig.~\ref{fig:listening_syngcn_all_left_pairs} reports the \% of ROI voxels with significant $R^2$ scores (hypothesis test where the difference in $R^2$ scores between the two feature groups is larger than 0) across feature groups for different ROIs in the left and right hemispheres, respectively.

We make the following observations from Fig.~\ref{fig:listening_syngcn_all_left_pairs}. (i) Unlike~\cite{reddy2021can}, we find that punctuation features yield a lower predictive performance across language regions for listening in both the left and right hemispheres. This is intuitive since punctuation marks are not ``visible'' when listening. (ii) Amongst CC, CI, INC, and DEP, after controlling for basic syntactic features \{PD, CM, PU\}, CC displays a large \% of significant voxels across multiple language sub-regions, largest in ATL, PTL, and MFG in left and in IFGOrb, PCC and dmPFC in the right hemispheres. This means there are voxels in these language sub-regions that capture hierarchical English grammar syntax beyond simple syntax signals captured by PD, CM, and PU. (iii) %CC and CI are obtained from constituency parse, INC from incremental top-down parse, and DEP from dependency parse. 
DEP parser explains addition variance after controlling for basic syntactic features for the AG region which is mainly a knowledge store of thematic relations between entities.
Also, DEP yields a large \% of significant voxels for the IFG region in the left hemisphere whereas PCC region in the right hemisphere.
Although INC does not show any additional variance in the left hemisphere, it performs well for IFG and MFG in the right hemisphere. (iv) On top of these representations, BERT adds to the variance the most in the context of CC, CI, INC, and DEP features in both hemispheres. 
%(vi) The \% of ROI voxels with significant increases is much higher in the left hemisphere compared to the right hemisphere, 
%(vii) The PD features have a lower \% of ROI voxels in the right hemisphere than the left hemisphere across all the language regions, and (viii) CM have greater variance than PD in right hemisphere compared to the left hemisphere.

%Overall, in summary, brain process linguistic structure information higher in language left hemisphere than right hemisphere across language sub regions.  
%Similar,~\cite{reddy2021can} concluded that BERT features after eliminating syntactic information yield higher percentage of voxels which are signification across language sub rois. It means that the voxels which are predicted using BERT are corresponding to more semantic and these voxels are distributed across language system.

%Figure~\ref{fig:listening_syngcn_semgcn_bert_r2} showcases the \% of significant voxels whose $R^2$ values are higher than chance for different language sub-regions for different feature groups. The observations from Figure~\ref{fig:listening_syngcn_semgcn_bert_r2} that the syntactic structure is distributed across different language regions. Also, there are more syntactic voxels which are significantly different from semantic voxels across the language sub regions.

\begin{figure*}[t] 
\centering
%\vspace{-1cm}
\includegraphics[width=0.8\linewidth]{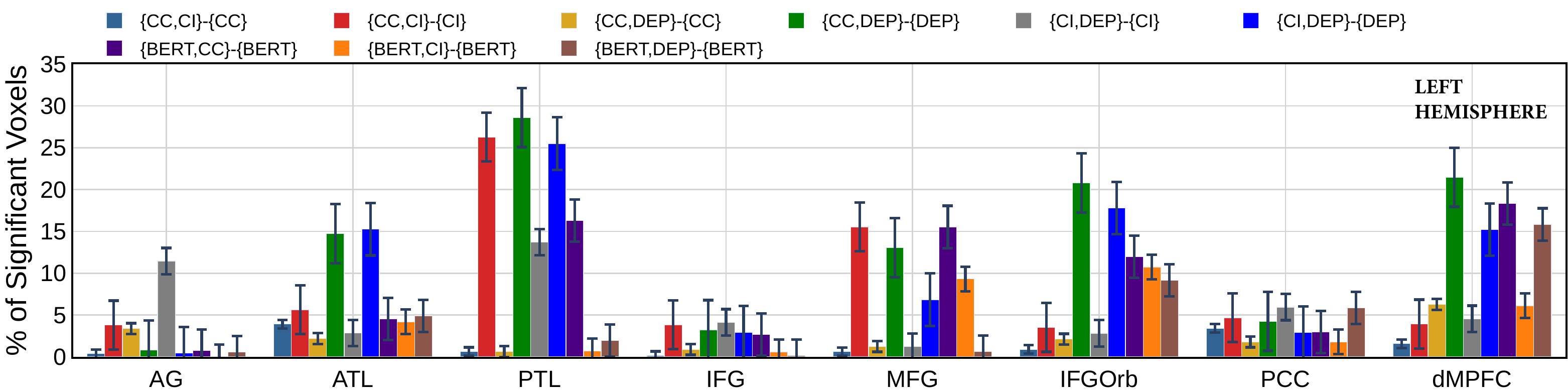}
%\vspace{-0.4cm}
\includegraphics[width=0.8\linewidth]{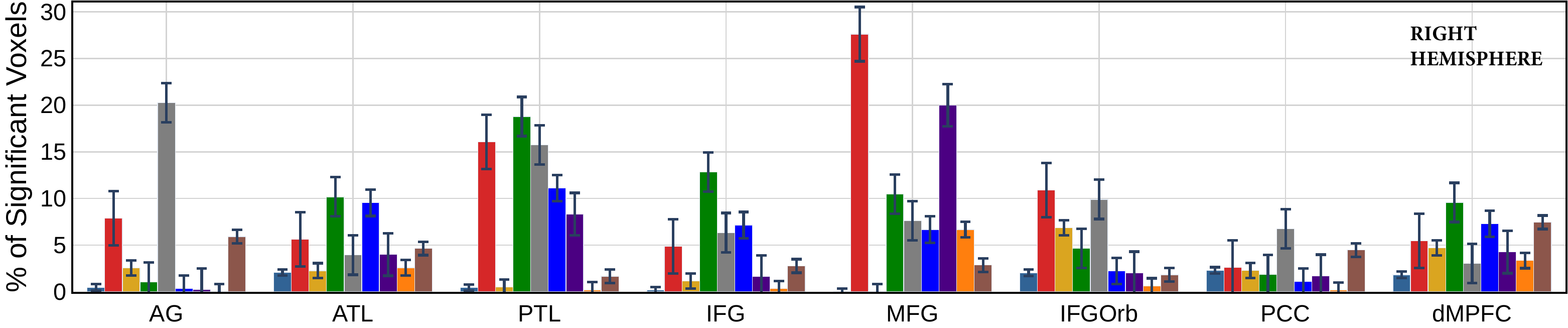}
\vspace{-0.3cm}
\caption{\textbf{Pairwise Predictive Power Comparison for Syntactic Parse Methods and BERT}: For each model, we show the percentage of ROI voxels in which we see a significant increase in prediction performance. Each bar represents the average percentage across 82 subjects, and the error bars show the standard error across subjects. ‘-’ indicates a hypothesis test for the difference in $R^2$ scores between the two feature groups being larger than 0. Left hemisphere (Top) and Right hemisphere (Bottom).}
\label{fig:listening_syngcn_bert_r2_left_new}
\end{figure*}

% \iffalse

% \begin{figure*}[!htb] 
% \centering
% \includegraphics[width=\linewidth]{images/narratives_syngcn_mainpairs_righthemisphere.pdf}
% \caption{Right-Hemisphere Region of Interest (ROI) analysis of the prediction performance for pairs of syntactic parsing feature sets and BERT. For each model, we show the percentage of ROI voxels in which we see significant increase in prediction performance. Each bar represents the average percentage across 82 subjects and the error bars show the standard error across subjects. ‘-’ indicates a hypothesis test for the difference in $R^2$ scores between the two feature groups being larger than 0.}
% \label{fig:listening_syngcn_bert_r2_right}
% \end{figure*}
% \fi

%\vspace{-2cm}

\noindent\textbf{Pairwise predictive power comparison for syntactic parse methods and BERT}
To compare relative extra syntactic information in various parse-based representations, we compute the difference in $R^2$ between every pair of representations from \{CC, CI, DEP\}. For this analysis, we ignore INC since it performed worst, as shown in Fig.~\ref{fig:listening_syngcn_all_left}.  Thus, we plot \% of significant ROI voxels for \{CC, DEP\}-\{CC\} and other such feature-pairwise combinations in Fig.~\ref{fig:listening_syngcn_bert_r2_left_new} for both hemispheres. We make the following observations from Fig.~\ref{fig:listening_syngcn_bert_r2_left_new}. (i) CC and CI show greater variance in brain predictivity (ATL and PTL for both hemispheres, MFG, IFGOrb and dmPFC of left hemisphere) even after controlling for either DEP. Also, CC and DEP show greater variance after controlling for CI. However, DEP or CI have negligible \% of ROI voxels after controlling for CC, specifically for temporal lobe (ATL and PTL) and frontal regions (IFG and MFG). Thus, we can conclude that constituency trees, specifically CC, encode similar syntactic information as DEP in temporal lobe (ATL and PTL) and frontal regions (IFG and MFG). Also, DEP based on dependency trees does not have additional syntactic information compared to constituency trees, except for AG, IFGOrb, PCC and dmPFC regions. 
%(ii) In comparison with INC, DEP have higher \% of ROI voxels after controlling INC where as INC have lower variation when controlling DEP. In particular, the variation is much higher in right hemisphere than left hemisphere. 
(ii) While BERT provides improvement over CC, CI and DEP in most brain areas (especially in MFG and dMPFC), surprisingly in AG and IFG, BERT does not provide much additive value.

\section{Discussion}
\vspace{-0.1cm}
% We studied three different parsing methods (constituency, dependency, and incremental top-down) along with basic syntactic feature representations and analyzed how effectively these representations predict the brain activity. 
% The central question pursued here is whether different regions of the language network mediate different syntactic computations. 
In this section, we correlate our empirical findings about syntactic parsing methods with previously proposed neuroscience theories. 
%We discuss analogy between the processing of syntax in the brain versus by parse trees across various aspects. 

% \noindent\textbf{How do brain vs. parse trees process syntax?}
%There is a broad agreement from psycholinguistic studies that semantics and syntax are specifically localized in Wernicke's and Broca's areas, respectively~\cite{grodzinsky2006neuroimaging}.
%Based on psycholinguistic studies,~\citet{grodzinsky2006neuroimaging} found that anterior temporal lobe (ATL) is involved in processing syntactic structure violations, implying that ATL is involved in identifying a mismatch between the incoming sentence and the expected syntactic structure of sentence. Thus it is plausible that during narrative listening, the human brain might be parsing the incoming sentences and predicting future structures that are appropriate in the context of the story.
From Fig.~\ref{fig:listening_syngcn_all_left_pairs}, we observe that activity in the left temporal lobe (ATL and PTL) seems to be predicted well using either CC or basic syntactic (PD) representations. These results are supported by theory of~\citet{matchin2020cortical}, who concluded that parts of the PTL are involved in hierarchical lexical-syntactic structure building, while the ATL is a knowledge store of entities.
While activity in the left IFGOrb, left PCC, and left AG seems to be better modeled by basic syntactic feature (PD) representations, that in MFG seems to be related to CC representations. DEP embeddings seem to perform better for activity in the left AG, left ATL and left IFG. This supports the theory of ~\citet{matchin2020cortical}, which reports that ATL is a knowledge store of entities and AG is a store of thematic relations between entities.  

A sub-ROI in the left AG, namely parietal area G inferior (PGi) has significantly more number of voxels sensitive to dependency features when we control for all other syntactic features. On the other hand sub-ROIs in the right temporo-parieto-occipital junction (TPOJ) are more sensitive to incremental top-down syntactic features (Appendix Fig.~\ref{fig:listening_ag_pairs}). While it is known that AG is sensitive to stimuli that are connected through a narrative rather than unconnected words~\cite{baker2018connectomic}, the current findings suggest that distinct sub-ROIs within AG are related to different syntactic features.  

Further sub-regions in the prefrontal cortex such as Brodmann area (BA) 44 and the inferior frontal junction area (IFJa) also seem to be related to representations of dependency parser (Appendix  Fig.~\ref{fig:listening_ifg_pairs}). The results in the prefrontal cortex seem to concur with the observations of~\citet{grodzinsky2006neuroimaging} and~\citet{kaan2002brain} who have shown that Broca’s area (Brodmann areas 44 and 45) has higher brain activation while processing complex sentences. Since narrative listening also involves processing highly complex sentences, consistent activation found in Left Brodmann areas 44 and 45 may relate to parsing of sentences or to see if they had distinct meanings.
%In particular, left posterior temporal lobe (PTL) support the integration of syntax and lexical information from complex sentences. Similarly, constituency complete trees and POS tags encode both lexical and syntactic information present in either simple or complex sentences.
The right hemisphere activation in the language network (AG, ATL, PTL, IFG, MFG, IFGOrb, PCC, and dMPFC) on the whole seems to be associated with basic syntactic features such as word length, word frequency, word count as embodied in CM representations.
%It is pointed out that syntactic operations are higher (i.e., ROIs of language network are highly active) for processing complex sentences than simple sentences~\cite{kaan2002brain}. 
%Since narrative listening also involves processing highly complex sentences, consistent activation found in Left Brodmann areas 44 and 45 may relate to parsing of sentences or to see if they had distinct meanings.
In linguistic studies, INC has been shown to be effective in checking if sentences with different syntax, have the same or different meaning. This in line with our observation that representations from INC parser seem to be more related to language regions (inferior frontal gyrus, IFG) in the right hemisphere as shown in Fig.~\ref{fig:listening_ifg_pairs}.

Overall,~\citet{grodzinsky2006neuroimaging} concluded that syntax processing is not limited to specific regions (left IFG or Broca's area). Along with IFG, other regions such as PTL, ATL, MFG, and IFGOrb are also involved in different stages of syntax processing. Our results (Fig.~\ref{fig:listening_syngcn_all_left}) also seem to support distributed representation of syntax across the language network. Moreover, our results clearly show the kind of syntax encoded by these individual ROIs.

%\vspace{-0.1cm}
\section{Conclusion}
\vspace{-0.2cm}
We studied the relative importance of multiple  constituency and dependency syntax parsing methods for fMRI prediction for the listening task. We find that (1) both CC and DEP are effective; CC is more important than CI,  (2) CC is better in temporal cortex and MFG,  while DEP is better in AG and PCC, (3) while BERT embeddings seem to be the most effective,  syntactic embedding methods also explain additional variance for a few ROIs. In line with previous works, we find that syntax and semantic processing is spread across multiple brain areas.

\section{Limitations}
Although these experiments were performed on only one dataset, it is indeed large with data from 82 participants. That said, it will be nice to perform experiments with more listening datasets.

We experiment with a linear encoder -- Ridge regression. 
%While some of the representations may have linear correlation with voxel magnitudes, non-linear transformations may be aligned with voxel magnitudes. 
We plan to experiment with more complex encoders as part of future work.

This work was done on data related to English stories only. Several other languages belong to the same language family as English~\cite{malik2022investigation}. While we can expect the insights and learnings to hold across languages in the same language family as English, empirical validation needs to be done. For languages in other language families, syntactic structure may be very different from English. Hence, more work needs to be done to check which of these insights hold for datasets in other language families. 

This work was conducted on a dataset where the participants were involved in the listening task. However, the stimuli was represented in the text form. We believe that an audio form of the stimuli can lead to improved insights. Thus, more work needs to be done to design representations (like prosodic features) for auditory stimuli.

\section{Ethical Statement}
We did not create any new fMRI data as part of this work. We used Narratives-Pieman dataset which is publicly available without any restrictions. Narratives dataset can be dowloaded from \url{https://datasets.datalad.org/?dir=/labs/hasson/narratives}. Please read their terms of use\footnote{\url{https://datasets.datalad.org/labs/hasson/narratives/stimuli/README}} for more details.

We do not foresee any harmful uses of this technology.

\bibliography{references}

\begin{thebibliography}{62}
\expandafter\ifx\csname natexlab\endcsname\relax\def\natexlab#1{#1}\fi

\bibitem[{Aw and Toneva(2022)}]{aw2022training}
Khai~Loong Aw and Mariya Toneva. 2022.
\newblock Training language models for deeper understanding improves brain
  alignment.
\newblock \emph{arXiv preprint arXiv:2212.10898}.

\bibitem[{Baker et~al.(2018)Baker, Burks, Briggs, Conner, Glenn, Taylor, Sali,
  McCoy, Battiste, O’Donoghue et~al.}]{baker2018connectomic}
Cordell~M Baker, Joshua~D Burks, Robert~G Briggs, Andrew~K Conner, Chad~A
  Glenn, Kathleen~N Taylor, Goksel Sali, Tressie~M McCoy, James~D Battiste,
  Daniel~L O’Donoghue, et~al. 2018.
\newblock A connectomic atlas of the human cerebrum—chapter 7: the lateral
  parietal lobe.
\newblock \emph{Operative Neurosurgery}, 15(suppl\_1):S295--S349.

\bibitem[{Bemis and Pylkk{\"a}nen(2011)}]{bemis2011simple}
Douglas~K Bemis and Liina Pylkk{\"a}nen. 2011.
\newblock Simple composition: A magnetoencephalography investigation into the
  comprehension of minimal linguistic phrases.
\newblock \emph{Journal of Neuroscience}, 31(8):2801--2814.

\bibitem[{Benjamini and Hochberg(1995)}]{benjamini1995controlling}
Yoav Benjamini and Yosef Hochberg. 1995.
\newblock Controlling the false discovery rate: a practical and powerful
  approach to multiple testing.
\newblock \emph{Journal of the Royal statistical society: series B
  (Methodological)}, 57(1):289--300.

\bibitem[{Bhattasali et~al.(2018)Bhattasali, Hale, Pallier, Brennan, Luh, and
  Spreng}]{bhattasali2018differentiating}
Shohini Bhattasali, John Hale, Christophe Pallier, Jonathan Brennan, Wen-Ming
  Luh, and R~Nathan Spreng. 2018.
\newblock Differentiating phrase structure parsing and memory retrieval in the
  brain.
\newblock \emph{Proceedings of the Society for Computation in Linguistics},
  1(1):74--80.

\bibitem[{Binder et~al.(2016)Binder, Conant, Humphries, Fernandino, Simons,
  Aguilar, and Desai}]{binder2016toward}
Jeffrey~R Binder, Lisa~L Conant, Colin~J Humphries, Leonardo Fernandino,
  Stephen~B Simons, Mario Aguilar, and Rutvik~H Desai. 2016.
\newblock Toward a brain-based componential semantic representation.
\newblock \emph{Cognitive neuropsychology}, 33(3-4):130--174.

\bibitem[{Blank et~al.(2016)Blank, Balewski, Mahowald, and
  Fedorenko}]{blank2016syntactic}
Idan Blank, Zuzanna Balewski, Kyle Mahowald, and Evelina Fedorenko. 2016.
\newblock Syntactic processing is distributed across the language system.
\newblock \emph{Neuroimage}, 127:307--323.

\bibitem[{Briscoe(1996)}]{briscoe1996syntax}
Ted Briscoe. 1996.
\newblock The syntax and semantics of punctuation and its use in
  interpretation.
\newblock In \emph{Proceedings of the Association for Computational Linguistics
  Workshop on Punctuation}, pages 1--7. Citeseer.

\bibitem[{Caramazza and Zurif(1976)}]{caramazza1976dissociation}
Alfonso Caramazza and Edgar~B Zurif. 1976.
\newblock Dissociation of algorithmic and heuristic processes in language
  comprehension: Evidence from aphasia.
\newblock \emph{Brain and language}, 3(4):572--582.

\bibitem[{Caucheteux et~al.(2021{\natexlab{a}})Caucheteux, Gramfort, and
  King}]{caucheteux2021disentangling}
Charlotte Caucheteux, Alexandre Gramfort, and Jean-Remi King.
  2021{\natexlab{a}}.
\newblock Disentangling syntax and semantics in the brain with deep networks.
\newblock In \emph{International Conference on Machine Learning}, pages
  1336--1348. PMLR.

\bibitem[{Caucheteux et~al.(2021{\natexlab{b}})Caucheteux, Gramfort, and
  King}]{caucheteux2021model}
Charlotte Caucheteux, Alexandre Gramfort, and Jean-R{\'e}mi King.
  2021{\natexlab{b}}.
\newblock Model-based analysis of brain activity reveals the hierarchy of
  language in 305 subjects.
\newblock \emph{arXiv preprint arXiv:2110.06078}.

\bibitem[{Cohen et~al.(2021)Cohen, Salondy, Pallier, and
  Dehaene}]{cohen2021does}
Laurent Cohen, Philippine Salondy, Christophe Pallier, and Stanislas Dehaene.
  2021.
\newblock How does inattention affect written and spoken language processing?
\newblock \emph{cortex}, 138:212--227.

\bibitem[{Desai et~al.(2022)Desai, Tadimeti, and Riccardi}]{desai2022proper}
Rutvik Desai, Usha Tadimeti, and Nicholas Riccardi. 2022.
\newblock Proper and common names in the semantic system.

\bibitem[{Devlin et~al.(2018)Devlin, Chang, Lee, and
  Toutanova}]{devlin2018bert}
Jacob Devlin, Ming-Wei Chang, Kenton Lee, and Kristina Toutanova. 2018.
\newblock Bert: Pre-training of deep bidirectional transformers for language
  understanding.
\newblock \emph{arXiv preprint arXiv:1810.04805}.

\bibitem[{Diakidoy et~al.(2005)Diakidoy, Stylianou, Karefillidou, and
  Papageorgiou}]{diakidoy2005relationship}
Irene-Anna~N Diakidoy, Polyxeni Stylianou, Christina Karefillidou, and
  Panayiota Papageorgiou. 2005.
\newblock The relationship between listening and reading comprehension of
  different types of text at increasing grade levels.
\newblock \emph{Reading psychology}, 26(1):55--80.

\bibitem[{Fedorenko et~al.(2020)Fedorenko, Blank, Siegelman, and
  Mineroff}]{fedorenko2020lack}
Evelina Fedorenko, Idan~Asher Blank, Matthew Siegelman, and Zachary Mineroff.
  2020.
\newblock Lack of selectivity for syntax relative to word meanings throughout
  the language network.
\newblock \emph{Cognition}, 203:104348.

\bibitem[{Fedorenko et~al.(2012)Fedorenko, Nieto-Castanon, and
  Kanwisher}]{fedorenko2012lexical}
Evelina Fedorenko, Alfonso Nieto-Castanon, and Nancy Kanwisher. 2012.
\newblock Lexical and syntactic representations in the brain: an fmri
  investigation with multi-voxel pattern analyses.
\newblock \emph{Neuropsychologia}, 50(4):499--513.

\bibitem[{Fedorenko and Thompson-Schill(2014)}]{fedorenko2014reworking}
Evelina Fedorenko and Sharon~L Thompson-Schill. 2014.
\newblock Reworking the language network.
\newblock \emph{Trends in cognitive sciences}, 18(3):120--126.

\bibitem[{Friederici(2011)}]{friederici2011brain}
Angela~D Friederici. 2011.
\newblock The brain basis of language processing: from structure to function.
\newblock \emph{Physiological reviews}, 91(4):1357--1392.

\bibitem[{Friederici et~al.(2006)Friederici, Fiebach, Schlesewsky, Bornkessel,
  and Von~Cramon}]{friederici2006processing}
Angela~D Friederici, Christian~J Fiebach, Matthias Schlesewsky, Ina~D
  Bornkessel, and D~Yves Von~Cramon. 2006.
\newblock Processing linguistic complexity and grammaticality in the left
  frontal cortex.
\newblock \emph{Cerebral Cortex}, 16(12):1709--1717.

\bibitem[{Genovese(2000)}]{genovese2000bayesian}
Christopher~R Genovese. 2000.
\newblock A bayesian time-course model for functional magnetic resonance
  imaging data.
\newblock \emph{Journal of the American Statistical Association},
  95(451):691--703.

\bibitem[{Glasser et~al.(2016)Glasser, Coalson, Robinson, Hacker, Harwell,
  Yacoub, Ugurbil, Andersson, Beckmann, Jenkinson et~al.}]{glasser2016multi}
Matthew~F Glasser, Timothy~S Coalson, Emma~C Robinson, Carl~D Hacker, John
  Harwell, Essa Yacoub, Kamil Ugurbil, Jesper Andersson, Christian~F Beckmann,
  Mark Jenkinson, et~al. 2016.
\newblock A multi-modal parcellation of human cerebral cortex.
\newblock \emph{Nature}, 536(7615):171--178.

\bibitem[{Grodzinsky and Friederici(2006)}]{grodzinsky2006neuroimaging}
Yosef Grodzinsky and Angela~D Friederici. 2006.
\newblock Neuroimaging of syntax and syntactic processing.
\newblock \emph{Current opinion in neurobiology}, 16(2):240--246.

\bibitem[{Handjaras et~al.(2016)Handjaras, Ricciardi, Leo, Lenci, Cecchetti,
  Cosottini, Marotta, and Pietrini}]{handjaras2016concepts}
Giacomo Handjaras, Emiliano Ricciardi, Andrea Leo, Alessandro Lenci, Luca
  Cecchetti, Mirco Cosottini, Giovanna Marotta, and Pietro Pietrini. 2016.
\newblock How concepts are encoded in the human brain: a modality independent,
  category-based cortical organization of semantic knowledge.
\newblock \emph{Neuroimage}, 135:232--242.

\bibitem[{Hays(1964)}]{hays1964dependency}
David~G Hays. 1964.
\newblock Dependency theory: A formalism and some observations.
\newblock \emph{Language}, 40(4):511--525.

\bibitem[{Hirst(1984)}]{hirst1984semantic}
Graeme Hirst. 1984.
\newblock A semantic process for syntactic disambiguation.
\newblock In \emph{AAAI}, pages 148--152.

\bibitem[{Hollenstein et~al.(2019)Hollenstein, de~la Torre, Langer, and
  Zhang}]{hollenstein2019cognival}
Nora Hollenstein, A~de~la Torre, Nicolas Langer, and Ce~Zhang. 2019.
\newblock Cognival: A framework for cognitive word embedding evaluation.
\newblock In \emph{Proceedings of The SIGNLL Conference on Computational
  Natural Language Learning 2019}.

\bibitem[{Honnibal and Montani(2017)}]{spacy2}
Matthew Honnibal and Ines Montani. 2017.
\newblock {spaCy 2}: Natural language understanding with {B}loom embeddings,
  convolutional neural networks and incremental parsing.
\newblock To appear.

\bibitem[{Humphries et~al.(2006)Humphries, Binder, Medler, and
  Liebenthal}]{humphries2006syntactic}
Colin Humphries, Jeffrey~R Binder, David~A Medler, and Einat Liebenthal. 2006.
\newblock Syntactic and semantic modulation of neural activity during auditory
  sentence comprehension.
\newblock \emph{Journal of cognitive neuroscience}, 18(4):665--679.

\bibitem[{Jain and Huth(2018)}]{jain2018incorporating}
Shailee Jain and Alexander~G Huth. 2018.
\newblock Incorporating context into language encoding models for fmri.
\newblock In \emph{Proceedings of the 32nd International Conference on Neural
  Information Processing Systems}, pages 6629--6638.

\bibitem[{Jawahar et~al.(2019)Jawahar, Sagot, and Seddah}]{jawahar2019does}
Ganesh Jawahar, Beno{\^\i}t Sagot, and Djam{\'e} Seddah. 2019.
\newblock What does bert learn about the structure of language?
\newblock In \emph{ACL 2019-57th Annual Meeting of the Association for
  Computational Linguistics}.

\bibitem[{Jung(1995)}]{jung1995syntaktische}
Wha-Young Jung. 1995.
\newblock \emph{Syntaktische Relationen im Rahmen der Dependenzgrammatik},
  volume~9.
\newblock Buske Verlag.

\bibitem[{Kaan and Swaab(2002)}]{kaan2002brain}
Edith Kaan and Tamara~Y Swaab. 2002.
\newblock The brain circuitry of syntactic comprehension.
\newblock \emph{Trends in cognitive sciences}, 6(8):350--356.

\bibitem[{LeBel et~al.(2021)LeBel, Jain, and Huth}]{lebel2021voxelwise}
Amanda LeBel, Shailee Jain, and Alexander~G Huth. 2021.
\newblock Voxelwise encoding models show that cerebellar language
  representations are highly conceptual.
\newblock \emph{Journal of Neuroscience}, 41(50):10341--10355.

\bibitem[{Luoma and Pyysalo(2020)}]{luoma2020exploring}
Jouni Luoma and Sampo Pyysalo. 2020.
\newblock Exploring cross-sentence contexts for named entity recognition with
  bert.
\newblock In \emph{Proceedings of the 28th International Conference on
  Computational Linguistics}, pages 904--914.

\bibitem[{Malik-Moraleda et~al.(2022)Malik-Moraleda, Ayyash, Gall{\'e}e,
  Affourtit, Hoffmann, Mineroff, Jouravlev, and
  Fedorenko}]{malik2022investigation}
Saima Malik-Moraleda, Dima Ayyash, Jeanne Gall{\'e}e, Josef Affourtit, Malte
  Hoffmann, Zachary Mineroff, Olessia Jouravlev, and Evelina Fedorenko. 2022.
\newblock An investigation across 45 languages and 12 language families reveals
  a universal language network.
\newblock \emph{Nature Neuroscience}, 25(8):1014--1019.

\bibitem[{Manning et~al.(2014)Manning, Surdeanu, Bauer, Finkel, Bethard, and
  McClosky}]{manning2014stanford}
Christopher~D Manning, Mihai Surdeanu, John Bauer, Jenny~Rose Finkel, Steven
  Bethard, and David McClosky. 2014.
\newblock The stanford corenlp natural language processing toolkit.
\newblock In \emph{Proceedings of 52nd annual meeting of the association for
  computational linguistics: system demonstrations}, pages 55--60.

\bibitem[{Matchin and Hickok(2020)}]{matchin2020cortical}
William Matchin and Gregory Hickok. 2020.
\newblock The cortical organization of syntax.
\newblock \emph{Cerebral Cortex}, 30(3):1481--1498.

\bibitem[{Merlin and Toneva(2022)}]{merlin2022language}
Gabriele Merlin and Mariya Toneva. 2022.
\newblock Language models and brain alignment: beyond word-level semantics and
  prediction.
\newblock \emph{arXiv preprint arXiv:2212.00596}.

\bibitem[{Milton et~al.(2021)Milton, Dhanaraj, Young, Taylor, Nicholas, Briggs,
  Bai, Fonseka, Hormovas, Lin et~al.}]{milton2021parcellation}
Camille~K Milton, Vukshitha Dhanaraj, Isabella~M Young, Hugh~M Taylor, Peter~J
  Nicholas, Robert~G Briggs, Michael~Y Bai, Rannulu~D Fonseka, Jorge Hormovas,
  Yueh-Hsin Lin, et~al. 2021.
\newblock Parcellation-based anatomic model of the semantic network.
\newblock \emph{Brain and behavior}, 11(4):e02065.

\bibitem[{Nastase et~al.(2021)Nastase, Liu, Hillman, Zadbood, Hasenfratz,
  Keshavarzian, Chen, Honey, Yeshurun, Regev et~al.}]{nastase2021narratives}
Samuel~A Nastase, Yun-Fei Liu, Hanna Hillman, Asieh Zadbood, Liat Hasenfratz,
  Neggin Keshavarzian, Janice Chen, Christopher~J Honey, Yaara Yeshurun, Mor
  Regev, et~al. 2021.
\newblock The “narratives” fmri dataset for evaluating models of
  naturalistic language comprehension.
\newblock \emph{Scientific data}, 8(1):1--22.

\bibitem[{Oota et~al.(2022{\natexlab{a}})Oota, Alexandre, and
  Hinaut}]{oota2022long}
Subba~Reddy Oota, Frederic Alexandre, and Xavier Hinaut. 2022{\natexlab{a}}.
\newblock Long-term plausibility of language models and neural dynamics during
  narrative listening.
\newblock In \emph{Proceedings of the Annual Meeting of the Cognitive Science
  Society}, volume~44.

\bibitem[{Oota et~al.(2022{\natexlab{b}})Oota, Arora, Agarwal, Marreddy, Gupta,
  and Surampudi}]{oota2022neural}
Subba~Reddy Oota, Jashn Arora, Veeral Agarwal, Mounika Marreddy, Manish Gupta,
  and Bapi~Raju Surampudi. 2022{\natexlab{b}}.
\newblock Neural language taskonomy: Which nlp tasks are the most predictive of
  fmri brain activity?
\newblock \emph{arXiv preprint arXiv:2205.01404}.

\bibitem[{Oota et~al.(2022{\natexlab{c}})Oota, Arora, Rowtula, Gupta, and
  Bapi}]{oota2022visio}
Subba~Reddy Oota, Jashn Arora, Vijay Rowtula, Manish Gupta, and Raju~S Bapi.
  2022{\natexlab{c}}.
\newblock Visio-linguistic brain encoding.
\newblock \emph{arXiv preprint arXiv:2204.08261}.

\bibitem[{Oota et~al.(2022{\natexlab{d}})Oota, Gupta, and
  Toneva}]{oota2022joint}
Subba~Reddy Oota, Manish Gupta, and Mariya Toneva. 2022{\natexlab{d}}.
\newblock Joint processing of linguistic properties in brains and language
  models.
\newblock \emph{arXiv preprint arXiv:2212.08094}.

\bibitem[{Oota et~al.(2018)Oota, Manwani, and Bapi}]{oota2018fmri}
Subba~Reddy Oota, Naresh Manwani, and Raju~S Bapi. 2018.
\newblock {fMRI Semantic Category Decoding Using Linguistic Encoding of Word
  Embeddings}.
\newblock In \emph{International Conference on Neural Information Processing},
  pages 3--15. Springer.

\bibitem[{Pallier et~al.(2011)Pallier, Devauchelle, and
  Dehaene}]{pallier2011cortical}
Christophe Pallier, Anne-Dominique Devauchelle, and Stanislas Dehaene. 2011.
\newblock Cortical representation of the constituent structure of sentences.
\newblock \emph{Proceedings of the National Academy of Sciences},
  108(6):2522--2527.

\bibitem[{Pennington et~al.(2014)Pennington, Socher, and
  Manning}]{pennington2014glove}
Jeffrey Pennington, Richard Socher, and Christopher~D Manning. 2014.
\newblock Glove: Global vectors for word representation.
\newblock In \emph{Proceedings of the 2014 conference on empirical methods in
  natural language processing (EMNLP)}, pages 1532--1543.

\bibitem[{Rambow(2010)}]{rambow2010simple}
Owen Rambow. 2010.
\newblock The simple truth about dependency and phrase structure
  representations: An opinion piece.
\newblock In \emph{Human language technologies: The 2010 annual conference of
  the North American Chapter of the Association for Computational Linguistics},
  pages 337--340.

\bibitem[{Reddy and Wehbe(2021)}]{reddy2021can}
Aniketh~Janardhan Reddy and Leila Wehbe. 2021.
\newblock Can fmri reveal the representation of syntactic structure in the
  brain?
\newblock \emph{Advances in Neural Information Processing Systems}, 34.

\bibitem[{Roark(2001)}]{roark2001probabilistic}
Brian Roark. 2001.
\newblock Probabilistic top-down parsing and language modeling.
\newblock \emph{Computational linguistics}, 27(2):249--276.

\bibitem[{Rogalsky and Hickok(2009)}]{rogalsky2009selective}
Corianne Rogalsky and Gregory Hickok. 2009.
\newblock Selective attention to semantic and syntactic features modulates
  sentence processing networks in anterior temporal cortex.
\newblock \emph{Cerebral Cortex}, 19(4):786--796.

\bibitem[{Rubin et~al.(2000)Rubin, Hafer, and Arata}]{rubin2000reading}
Donald~L Rubin, Teresa Hafer, and Kevin Arata. 2000.
\newblock Reading and listening to oral-based versus literate-based discourse.
\newblock \emph{Communication Education}, 49(2):121--133.

\bibitem[{Schneider(1998)}]{schneider1998linguistic}
Gerold Schneider. 1998.
\newblock \emph{A linguistic comparison of constituency, dependency and link
  grammar}.
\newblock Ph.D. thesis, Master’s thesis, University of Z{\"u}rich.

\bibitem[{Toneva et~al.(2022)Toneva, Mitchell, and Wehbe}]{toneva2022combining}
Mariya Toneva, Tom~M Mitchell, and Leila Wehbe. 2022.
\newblock Combining computational controls with natural text reveals aspects of
  meaning composition.
\newblock \emph{Nature Computational Science}, 2(11):745--757.

\bibitem[{Toneva et~al.(2020)Toneva, Stretcu, P{\'o}czos, Wehbe, and
  Mitchell}]{toneva2020modeling}
Mariya Toneva, Otilia Stretcu, Barnab{\'a}s P{\'o}czos, Leila Wehbe, and Tom~M
  Mitchell. 2020.
\newblock Modeling task effects on meaning representation in the brain via
  zero-shot meg prediction.
\newblock \emph{Advances in Neural Information Processing Systems},
  33:5284--5295.

\bibitem[{Toneva et~al.(2021)Toneva, Williams, Bollu, Dann, and
  Wehbe}]{toneva2021same}
Mariya Toneva, Jennifer Williams, Anand Bollu, Christoph Dann, and Leila Wehbe.
  2021.
\newblock Same cause; different effects in the brain.
\newblock In \emph{First Conference on Causal Learning and Reasoning}.

\bibitem[{Vashishth et~al.(2019)Vashishth, Bhandari, Yadav, Rai, Bhattacharyya,
  and Talukdar}]{vashishth2019incorporating}
Shikhar Vashishth, Manik Bhandari, Prateek Yadav, Piyush Rai, Chiranjib
  Bhattacharyya, and Partha~P Talukdar. 2019.
\newblock Incorporating syntactic and semantic information in word embeddings
  using graph convolutional networks.
\newblock In \emph{ACL (1)}.

\bibitem[{Wang et~al.(2020)Wang, Zhang, Lin, and Zong}]{wang2020probing}
Shaonan Wang, Jiajun Zhang, Nan Lin, and Chengqing Zong. 2020.
\newblock Probing brain activation patterns by dissociating semantics and
  syntax in sentences.
\newblock In \emph{Proceedings of the AAAI Conference on Artificial
  Intelligence}, volume~34, pages 9201--9208.

\bibitem[{Wehbe et~al.(2014)Wehbe, Murphy, Talukdar, Fyshe, Ramdas, and
  Mitchell}]{wehbe2014simultaneously}
Leila Wehbe, Brian Murphy, Partha Talukdar, Alona Fyshe, Aaditya Ramdas, and
  Tom Mitchell. 2014.
\newblock Simultaneously uncovering the patterns of brain regions involved in
  different story reading subprocesses.
\newblock \emph{PloS one}, 9(11):e112575.

\bibitem[{Zaccarella and Friederici(2015)}]{zaccarella2015merge}
Emiliano Zaccarella and Angela~D Friederici. 2015.
\newblock Merge in the human brain: A sub-region based functional investigation
  in the left pars opercularis.
\newblock \emph{Frontiers in psychology}, 6:1818.

\bibitem[{Zhang et~al.(2022)Zhang, Wang, Lin, Zhang, and
  Zong}]{zhang2022probing}
Xiaohan Zhang, Shaonan Wang, Nan Lin, Jiajun Zhang, and Chengqing Zong. 2022.
\newblock Probing word syntactic representations in the brain by a feature
  elimination method.
\newblock In \emph{AAAI}.

\end{thebibliography}

\appendix
\section{Hyper-parameter Settings}
\label{sec:hyperParameters}
All experiments were conducted on a machine with 1 NVIDIA GEFORCE-GTX GPU with 16GB GPU RAM. We used banded ridge-regression with following parameters: MSE loss function, and L2-decay ($\lambda$) varied from  10$^{-1}$ to 10$^{-3}$; best $\lambda$ was chosen by tuning on validation data; number of cross-validation runs was 4. 

\section{Constituency Complete Trees}
We now present the largest subtrees completed by a few of the words in the sentence: ``I began my illustrious career as a hard-boiled reporter in the Bronx where I toiled for the Ram, uh, Fordham University’s student newspaper''.

\begin{figure*}[t] 
\centering
\begin{minipage}{0.33\linewidth}
\centering
\includegraphics[width=0.2\linewidth, height=1.3in]{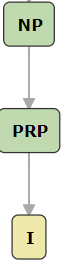} 
\caption*{(a) I}
\end{minipage}
\hspace{2cm}
\begin{minipage}{0.33\linewidth}
\centering
\includegraphics[width=0.2\linewidth,height=1.3in]{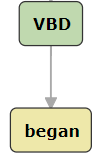}
\caption*{(b) began}
\end{minipage}
\begin{minipage}{\linewidth}
\centering
\includegraphics[width=\linewidth]{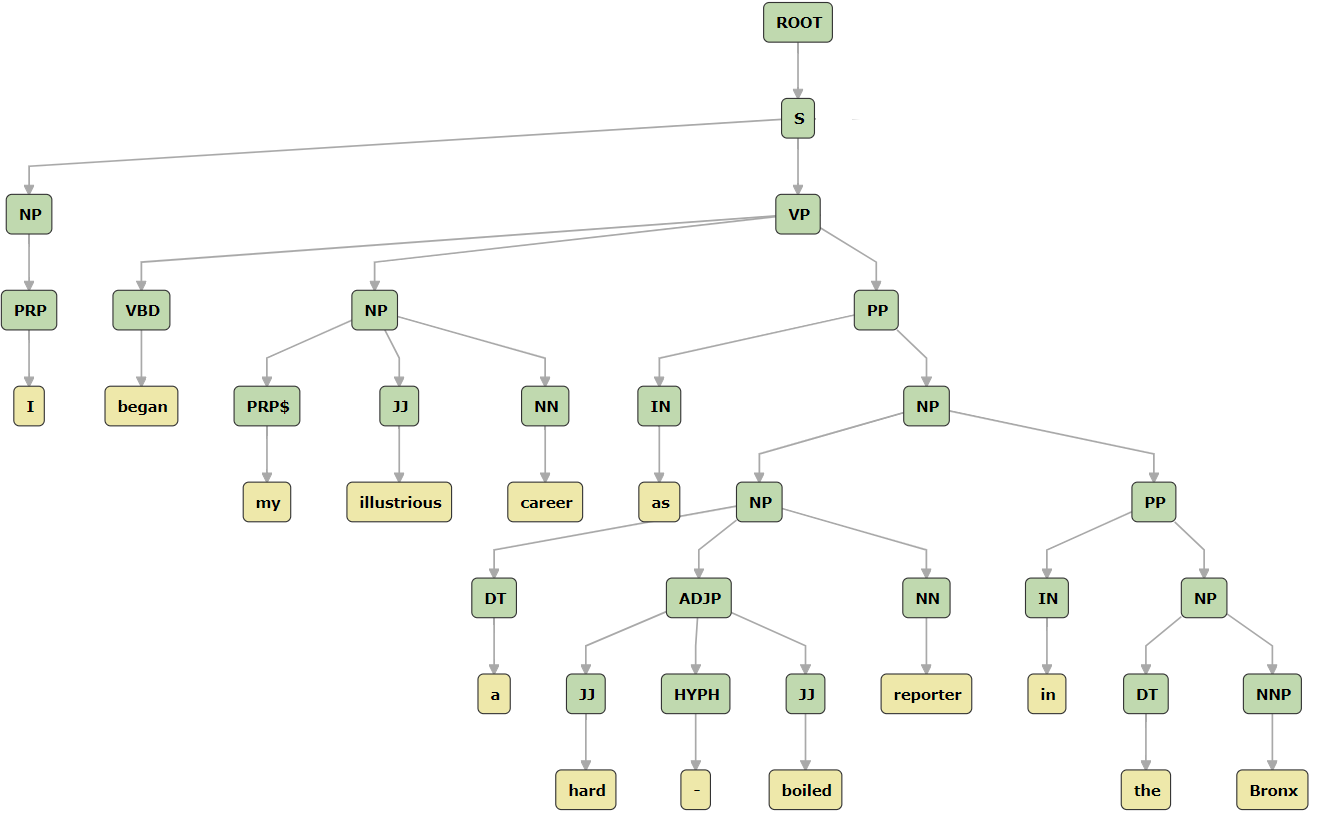}
\caption*{(c) Bronx}
\end{minipage}
\caption{Complete trees for the words: I, began, and Bronx,  for the sentence ``I began my illustrious career as a hard-boiled reporter in the Bronx where I toiled for the Ram, uh, Fordham University’s student newspaper.''}
\label{fig:complete_tree}
\end{figure*}

\begin{figure*}[t] 
\centering
\includegraphics[width=\linewidth]{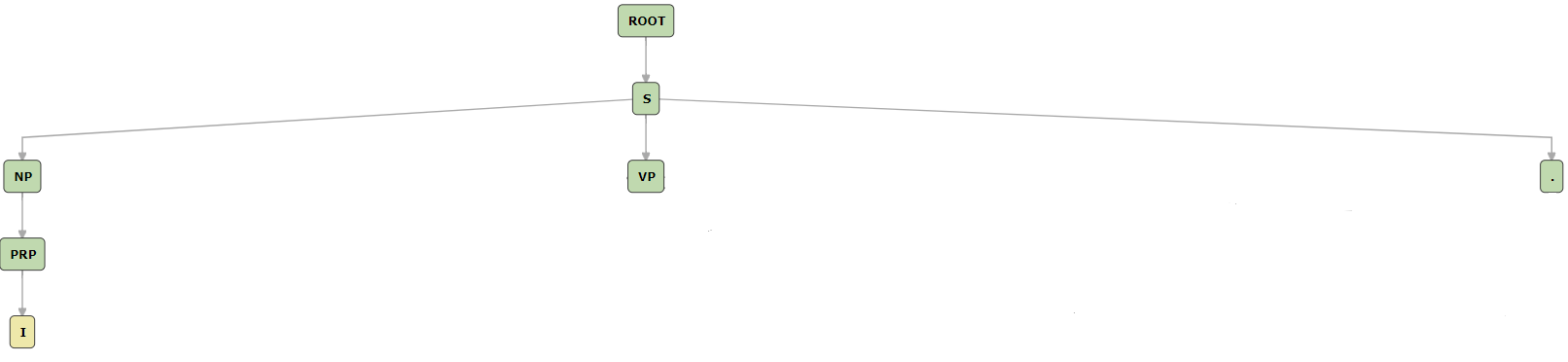}
\caption{Incomplete trees for the word: I, for the sentence ``I began my illustrious career as a hard-boiled reporter in the Bronx where I toiled for the Ram, uh, Fordham University’s student newspaper.''}
\label{fig:incomplete_tree}
\end{figure*}

\begin{figure*}[t] 
\centering
%\vspace{-1cm}
\includegraphics[width=0.8\linewidth]{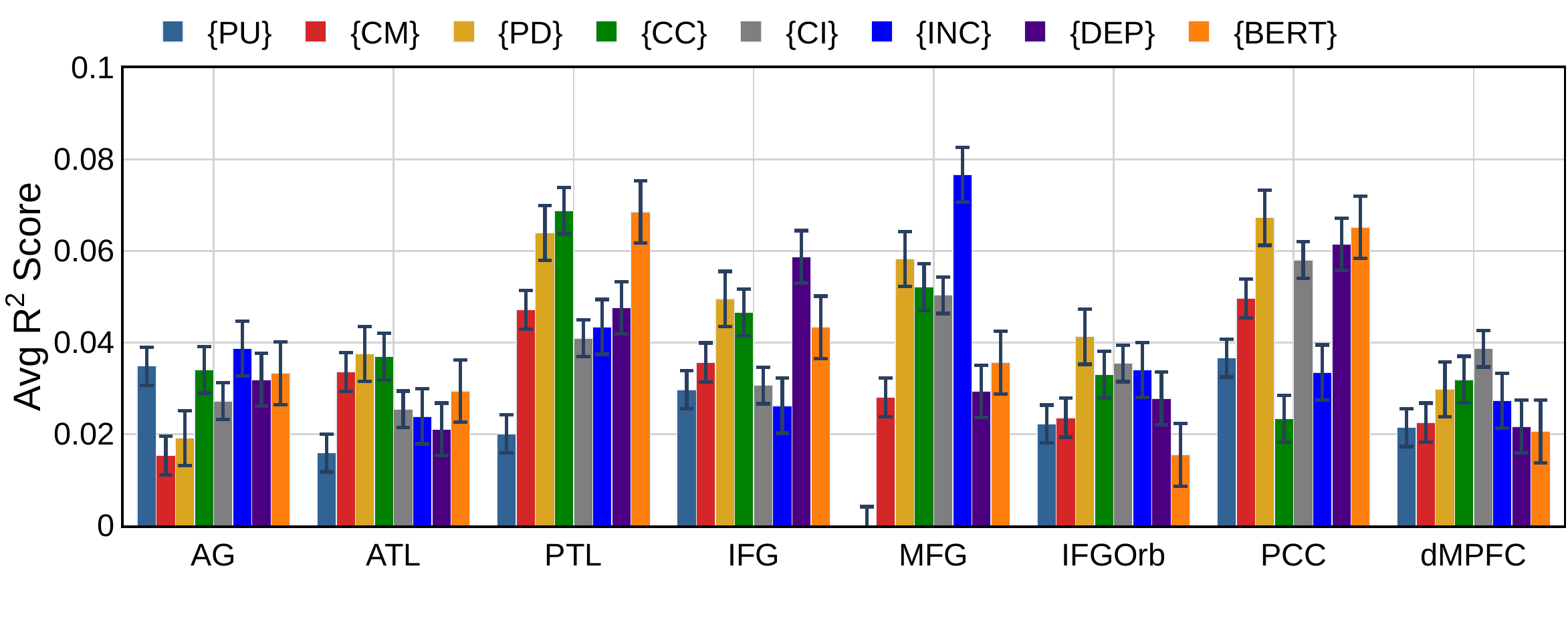}
%\vspace{-0.3cm}
\includegraphics[width=0.8\linewidth]{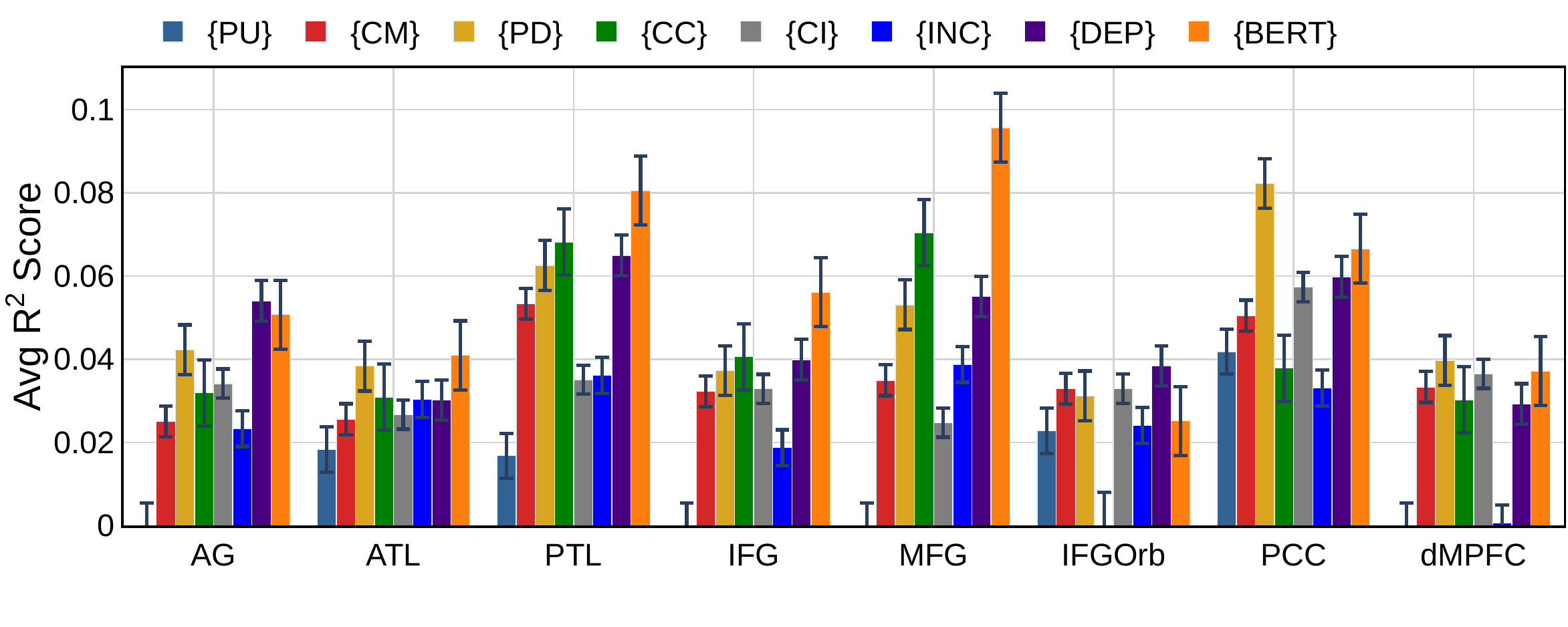}
%\vspace{-1cm}
\caption{\textbf{Performance of Individual Embedding Methods}: Region of Interest (ROI) analysis of the prediction performance  of various feature sets. For each model, we show $R^2$ score. Each bar represents the average score and error bars show standard error across 82 subjects. Left hemisphere (Top) and Right hemisphere (Bottom)}
%‘-’ indicates a hypothesis test for the difference in $R^2$ scores between the two feature groups being larger than 0.}
\label{fig:listening_avg_r2_allmodels}
\end{figure*}

\begin{figure*}[t] 
\centering
%\vspace{-1cm}
\includegraphics[width=0.85\linewidth]{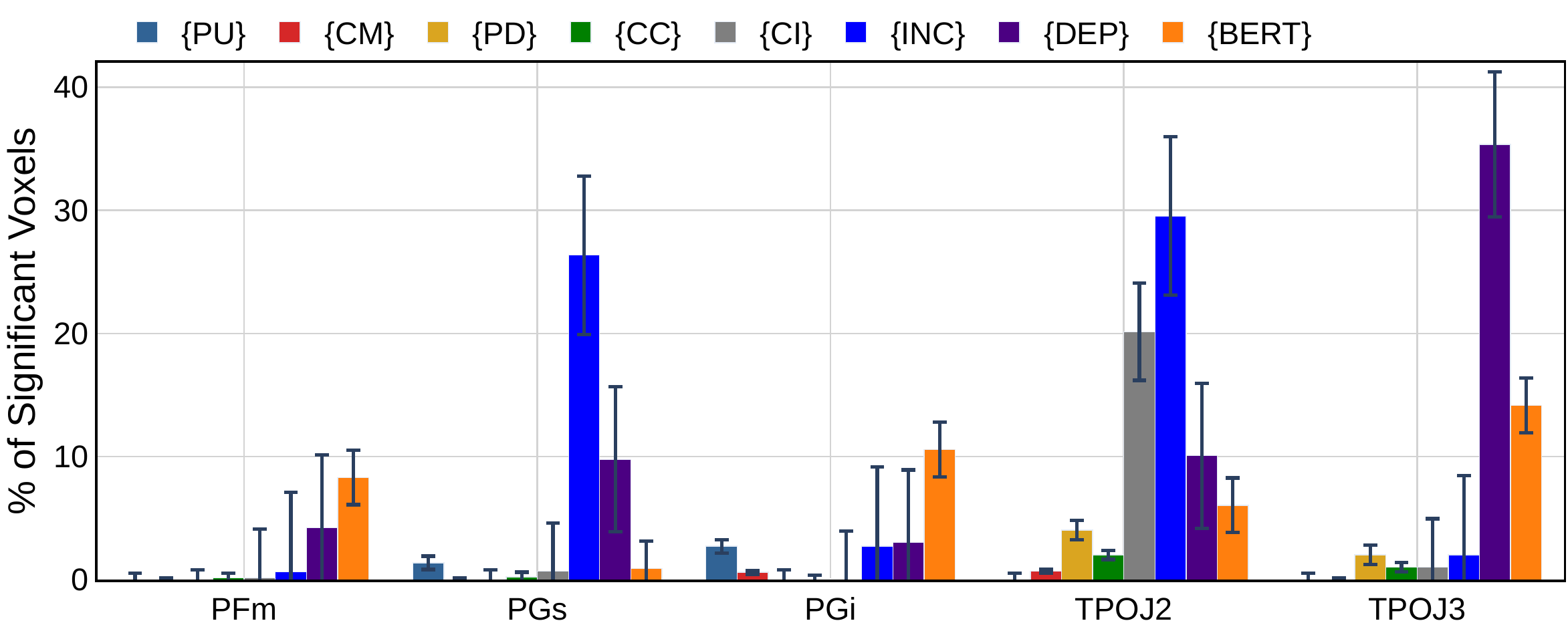}
%\vspace{-0.3cm}
\includegraphics[width=0.85\linewidth]{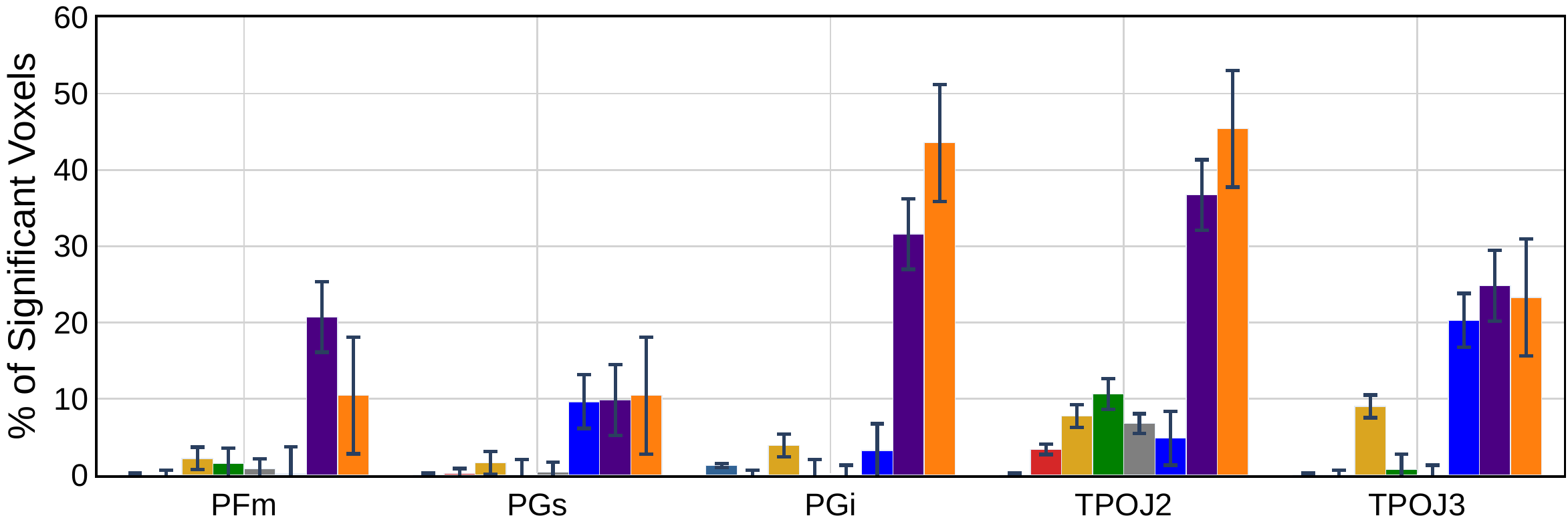}
%\vspace{-1cm}
\caption{\textbf{Performance of Individual Embedding Methods for AG region}: Region of Interest (ROI) analysis of the prediction performance  of various feature sets for various sub-regions. For each model, we show the \% of ROI voxels with a significant increase in prediction performance. Each bar represents the average score and error bars show standard error across 82 subjects. Left hemisphere (Top) and Right hemisphere (Bottom)}
%‘-’ indicates a hypothesis test for the difference in $R^2$ scores between the two feature groups being larger than 0.}
\label{fig:listening_avg_r2_AG}
\end{figure*}

\begin{figure*}[t] 
\centering
%\vspace{-1cm}
\includegraphics[width=0.85\linewidth]{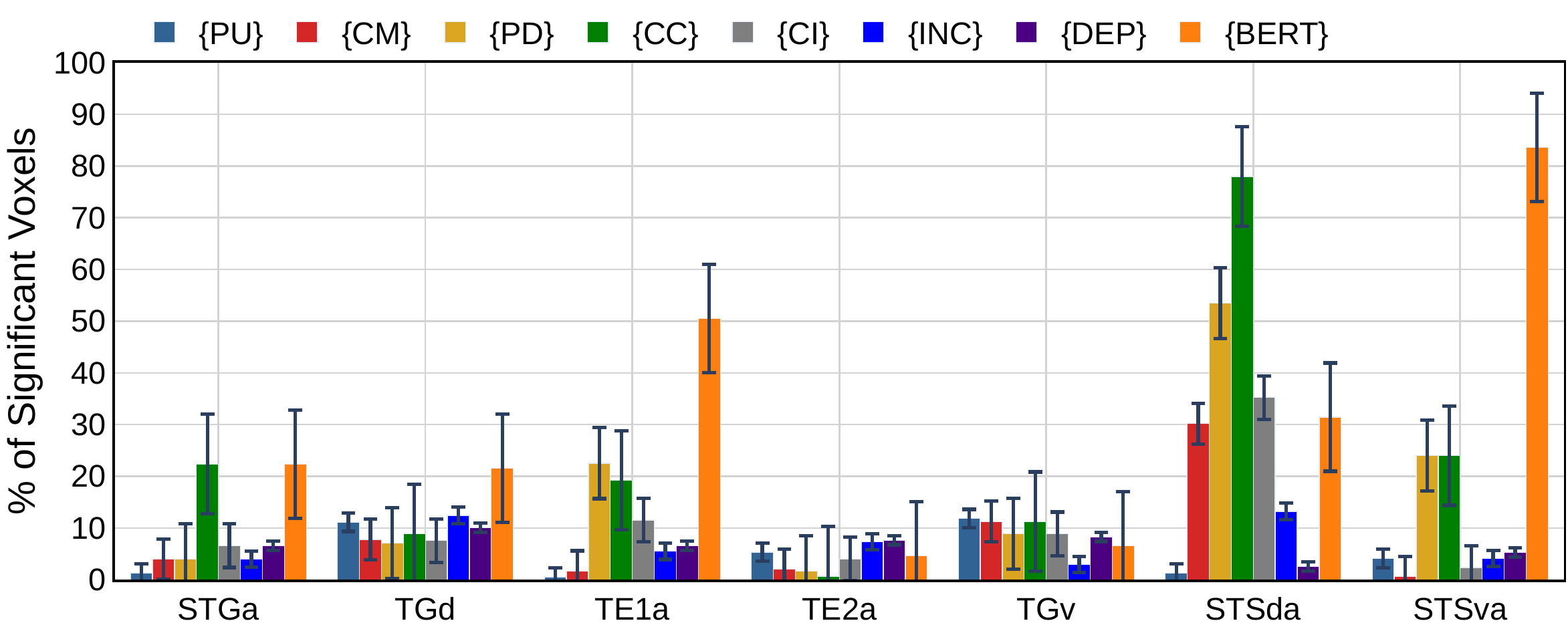}
%\vspace{-0.3cm}
\includegraphics[width=0.85\linewidth]{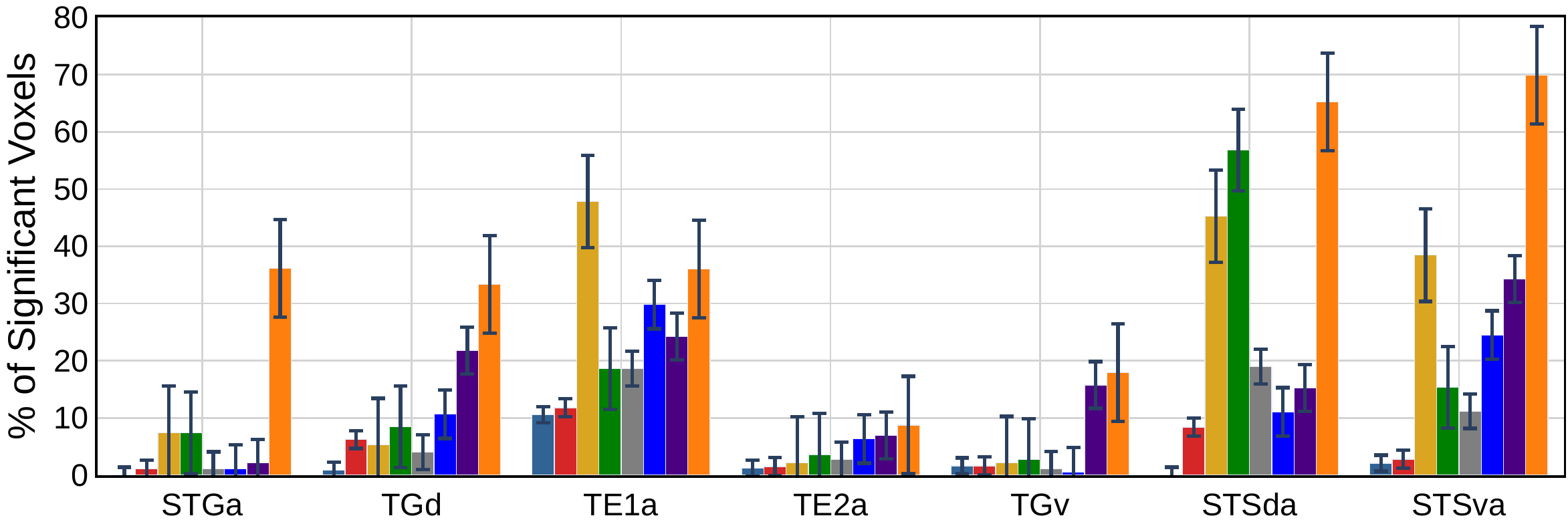}
%\vspace{-1cm}
\caption{\textbf{Performance of Individual Embedding Methods for ATL region}: Region of Interest (ROI) analysis of the prediction performance  of various feature sets for various sub-regions. For each model, we show the \% of ROI voxels with a significant increase in prediction performance. Each bar represents the average score and error bars show standard error across 82 subjects. Left hemisphere (Top) and Right hemisphere (Bottom)}
%‘-’ indicates a hypothesis test for the difference in $R^2$ scores between the two feature groups being larger than 0.}
\label{fig:listening_avg_r2_ATL}
\end{figure*}

\begin{figure*}[t] 
\centering
%\vspace{-1cm}
\includegraphics[width=0.85\linewidth]{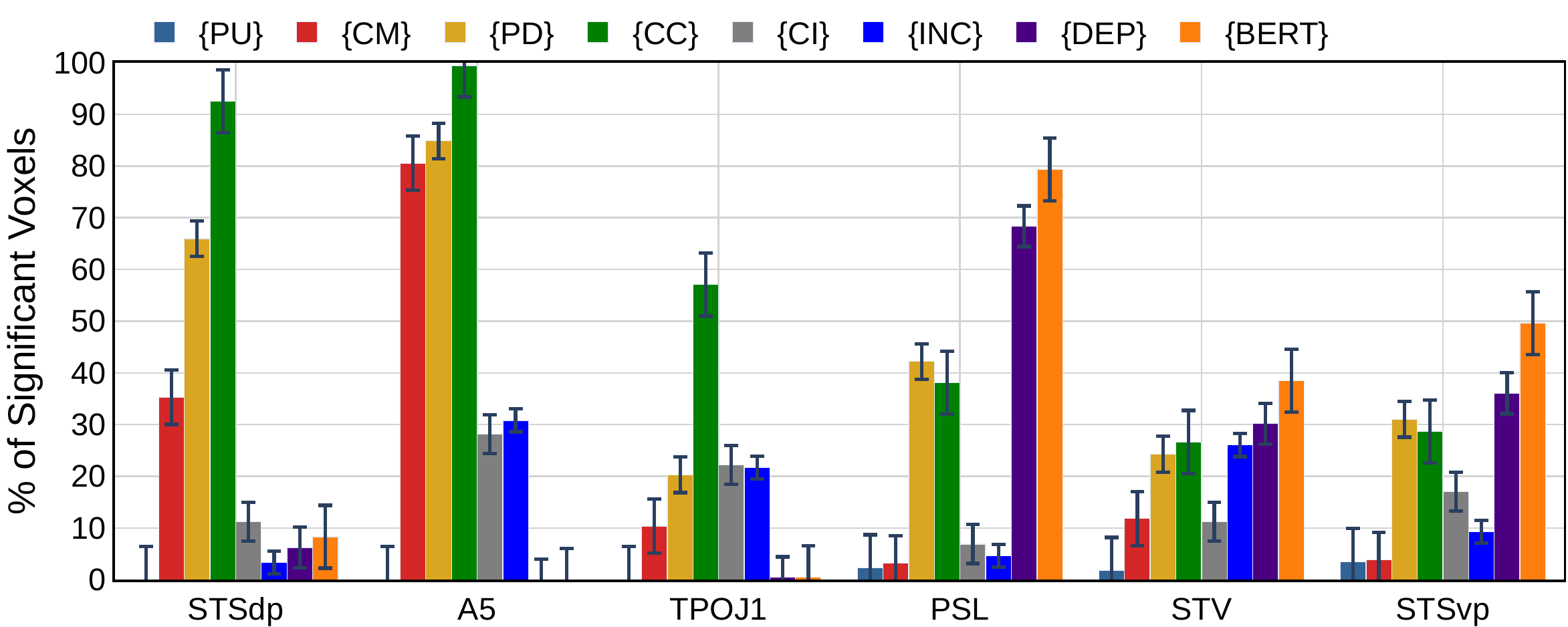}
%\vspace{-0.3cm}
\includegraphics[width=0.85\linewidth]{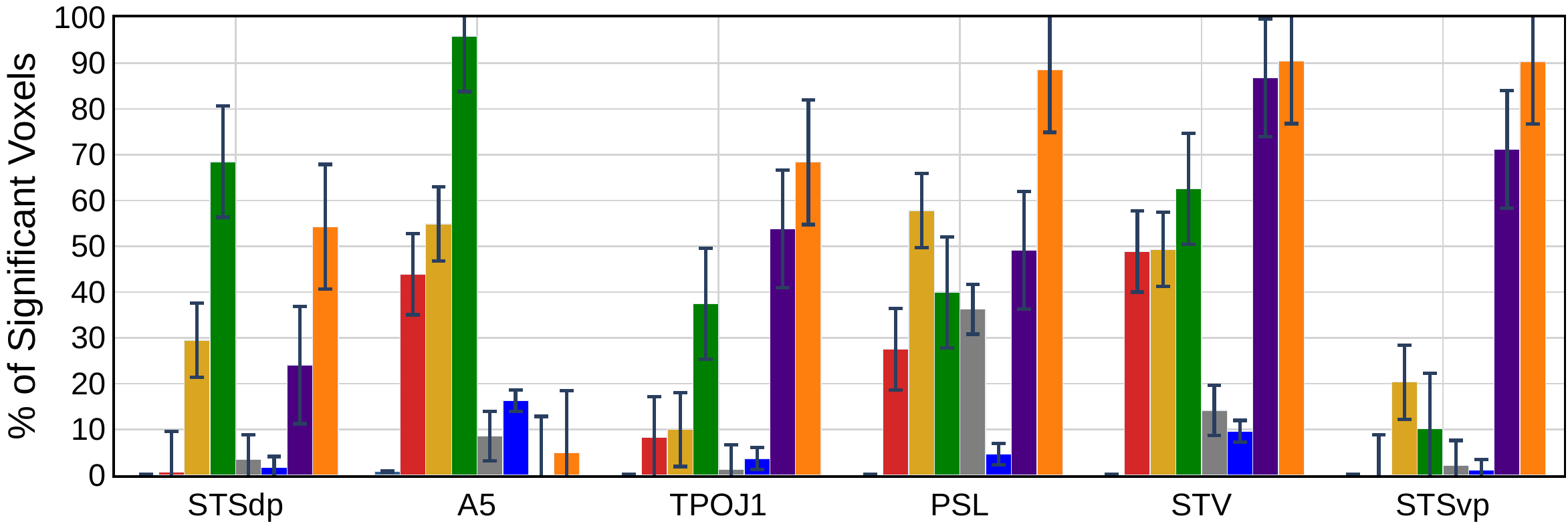}
%\vspace{-1cm}
\caption{\textbf{Performance of Individual Embedding Methods for PTL region}: Region of Interest (ROI) analysis of the prediction performance  of various feature sets for various sub-regions of PTL region. For each model, we show the \% of ROI voxels with a significant increase in prediction performance. Each bar represents the average score and error bars show standard error across 82 subjects. Left hemisphere (Top) and Right hemisphere (Bottom)}
%‘-’ indicates a hypothesis test for the difference in $R^2$ scores between the two feature groups being larger than 0.}
\label{fig:listening_avg_r2_PTL}
\end{figure*}

\begin{figure*}[t] 
\centering
%\vspace{-1cm}
\includegraphics[width=0.8\linewidth]{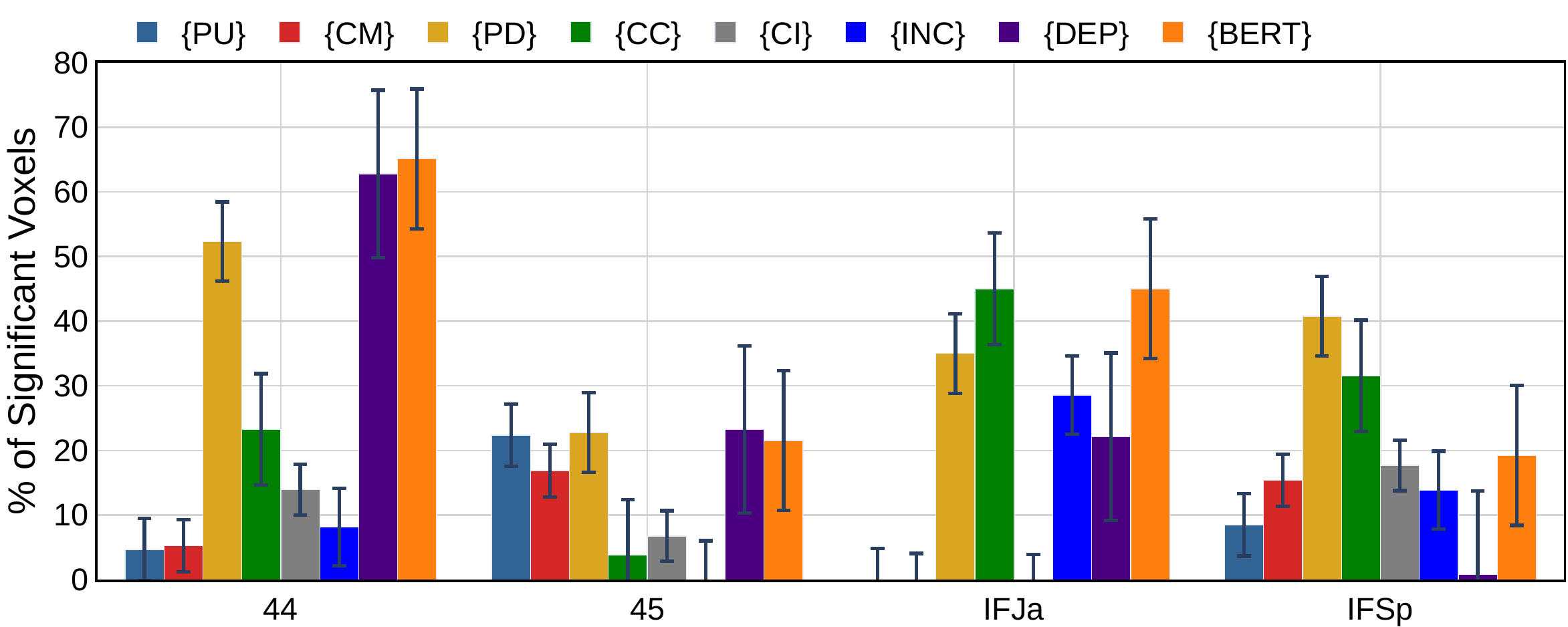}
%\vspace{-0.3cm}
\includegraphics[width=0.8\linewidth]{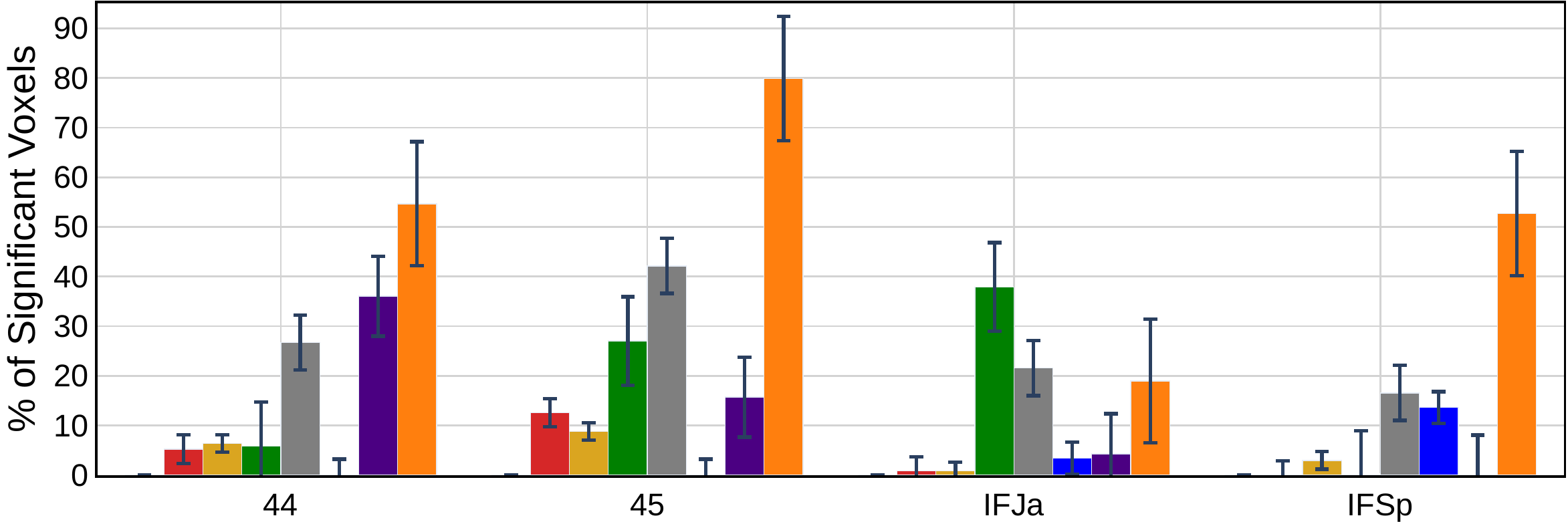}
%\vspace{-1cm}
\caption{\textbf{Performance of Individual Embedding Methods for IFG region}: Region of Interest (ROI) analysis of the prediction performance  of various feature sets for various sub-regions of IFG region. For each model, we show the \% of ROI voxels with a significant increase in prediction performance. Each bar represents the average score and error bars show standard error across 82 subjects. Left hemisphere (Top) and Right hemisphere (Bottom).}
%‘-’ indicates a hypothesis test for the difference in $R^2$ scores between the two feature groups being larger than 0.}
\label{fig:listening_avg_r2_ifg}
\end{figure*}

\begin{figure*}[t] 
\centering
%\vspace{-1cm}
\includegraphics[width=0.85\linewidth]{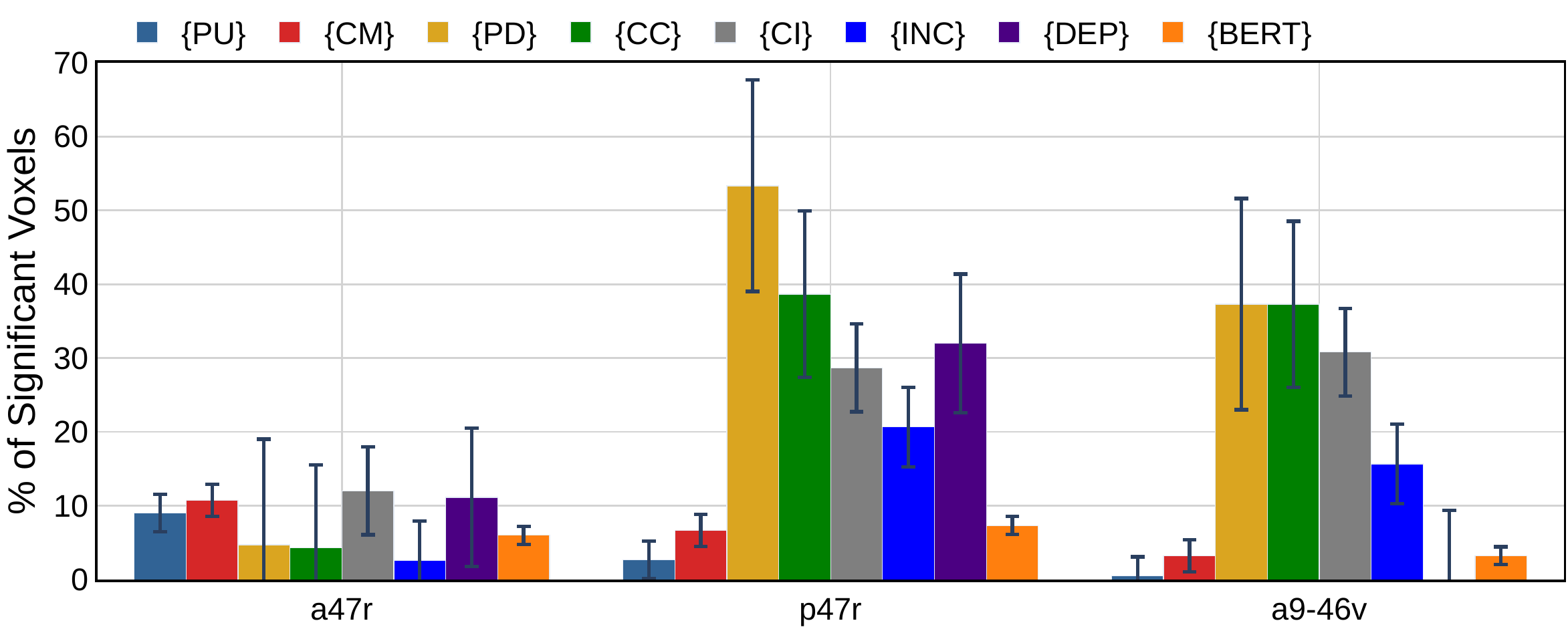}
%\vspace{-0.3cm}
\includegraphics[width=0.85\linewidth]{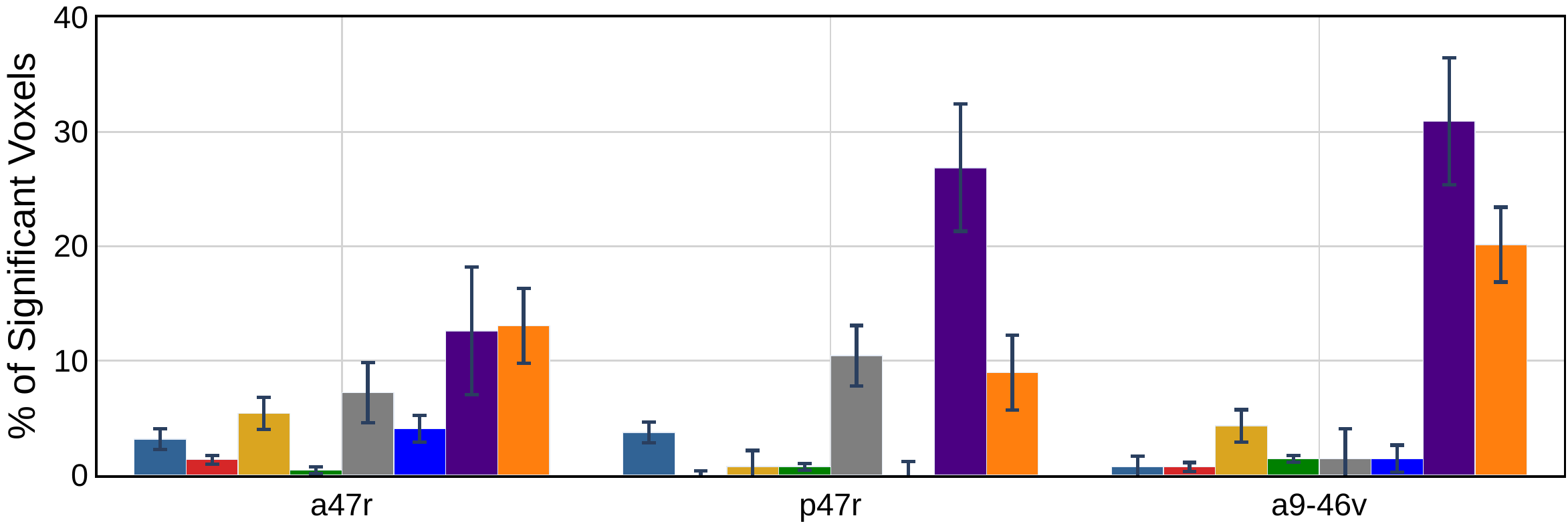}
%\vspace{-1cm}
\caption{\textbf{Performance of Individual Embedding Methods for IFGOrb region}: Region of Interest (ROI) analysis of the prediction performance  of various feature sets for various sub-regions of IFGOrb region. For each model, we show the \% of ROI voxels with a significant increase in prediction performance. Each bar represents the average score and error bars show standard error across 82 subjects. Left hemisphere (Top) and Right hemisphere (Bottom).}
%‘-’ indicates a hypothesis test for the difference in $R^2$ scores between the two feature groups being larger than 0.}
\label{fig:listening_avg_r2_ifgorb}
\end{figure*}

\begin{figure*}[t] 
\centering
%\vspace{-1cm}
\includegraphics[width=0.85\linewidth]{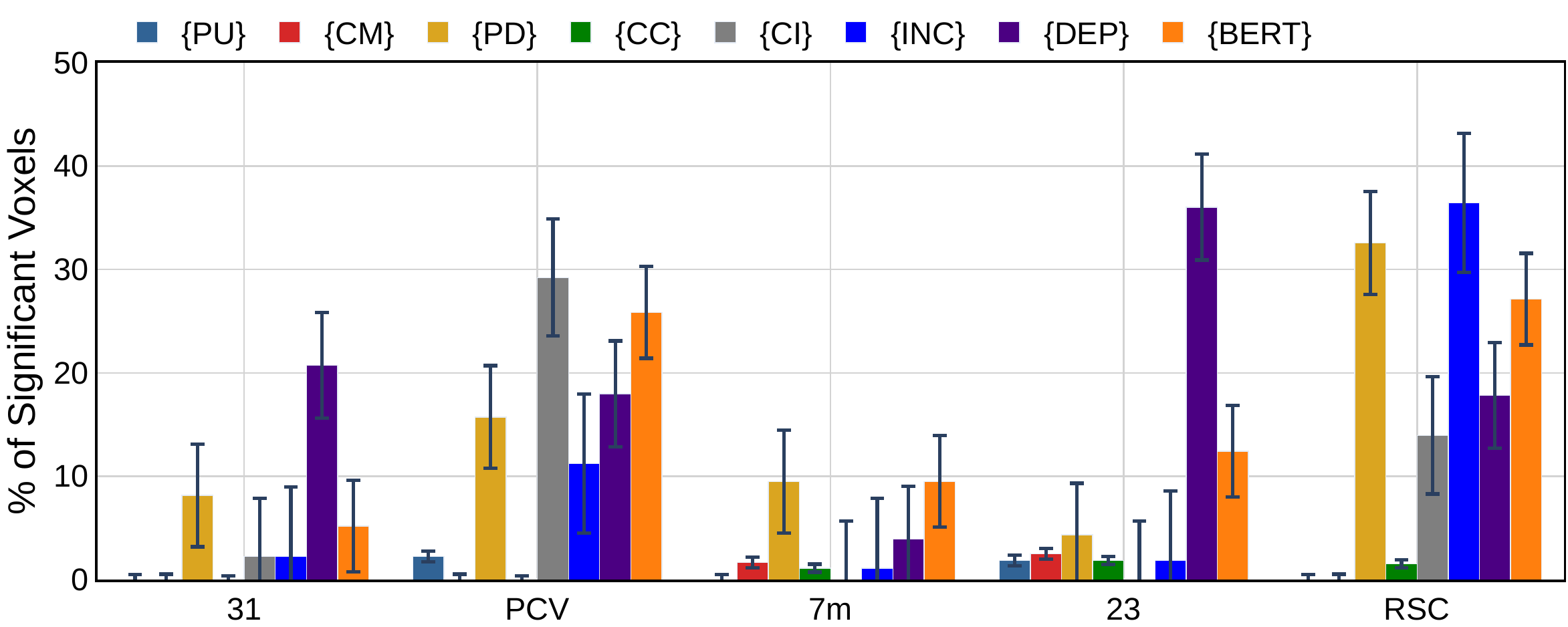}
%\vspace{-0.3cm}
\includegraphics[width=0.85\linewidth]{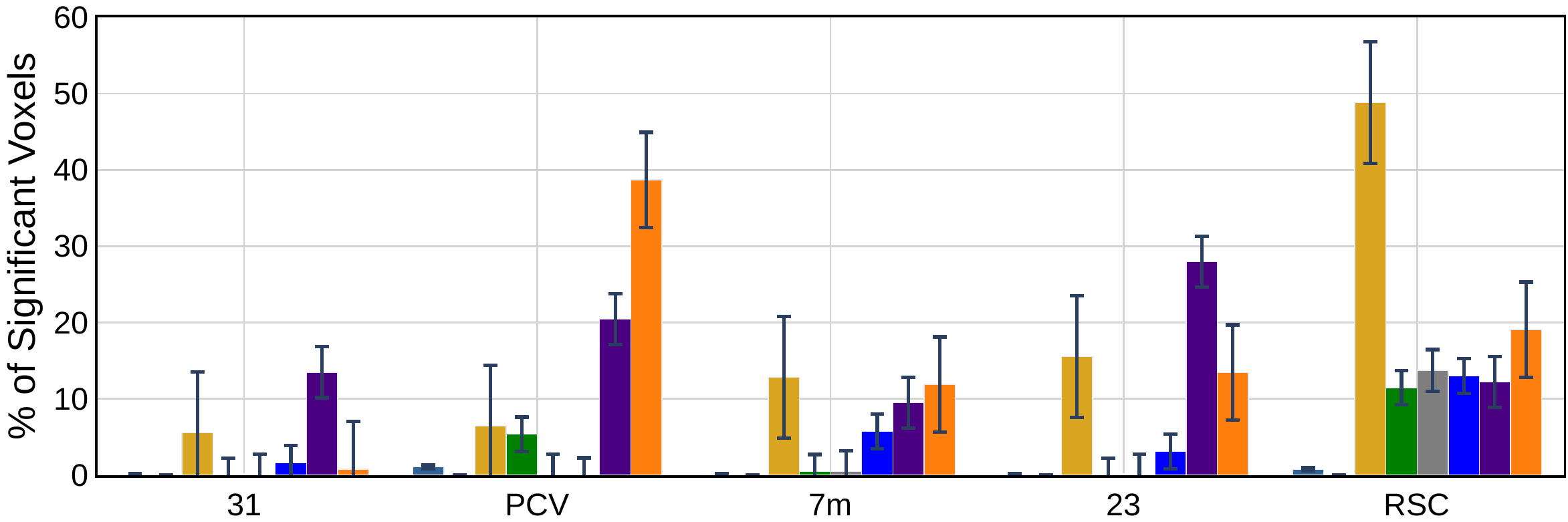}
%\vspace{-1cm}
\caption{\textbf{Performance of Individual Embedding Methods for PCC region}: Region of Interest (ROI) analysis of the prediction performance  of various feature sets for various sub-regions of PCC region. For each model, we show the \% of ROI voxels with a significant increase in prediction performance. Each bar represents the average score and error bars show standard error across 82 subjects. Left hemisphere (Top) and Right hemisphere (Bottom).}
%‘-’ indicates a hypothesis test for the difference in $R^2$ scores between the two feature groups being larger than 0.}
\label{fig:listening_avg_r2_pcc}
\end{figure*}

\begin{figure*}[t] 
\centering
%\vspace{-1cm}
\includegraphics[width=0.85\linewidth]{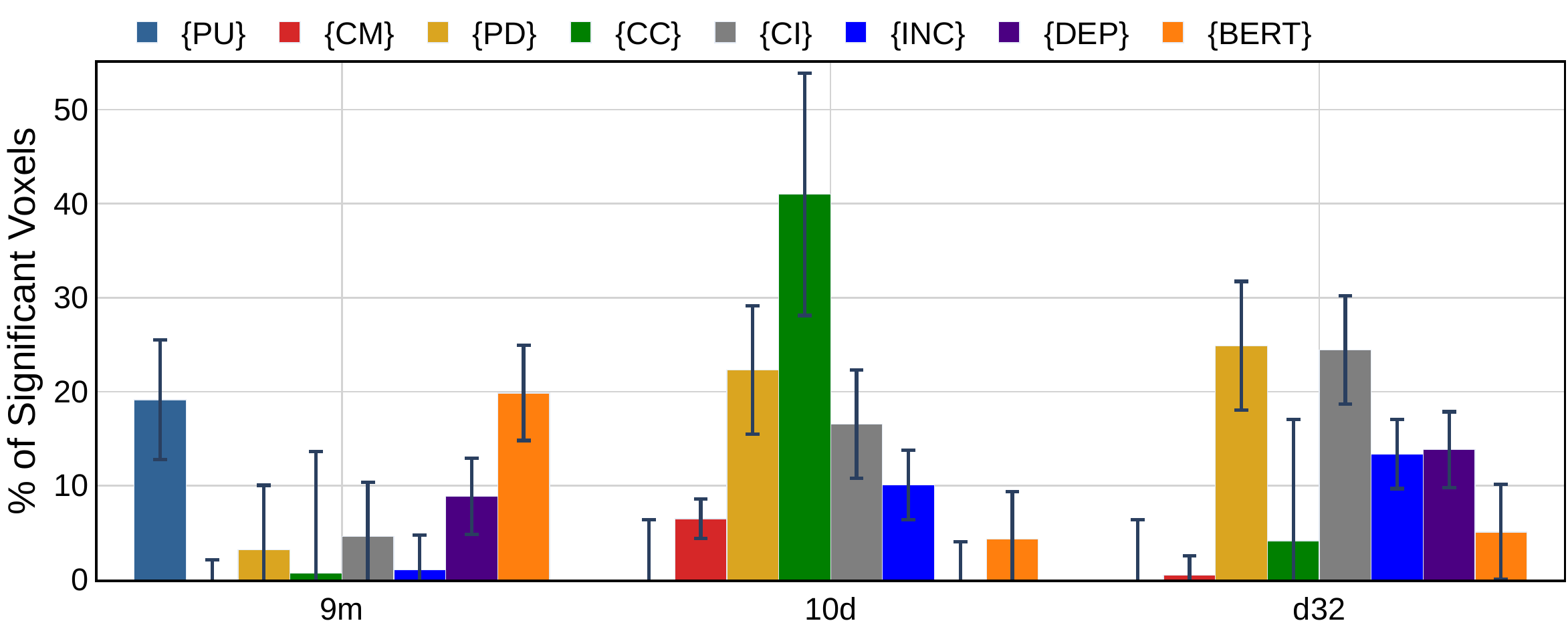}
%\vspace{-0.3cm}
\includegraphics[width=0.85\linewidth]{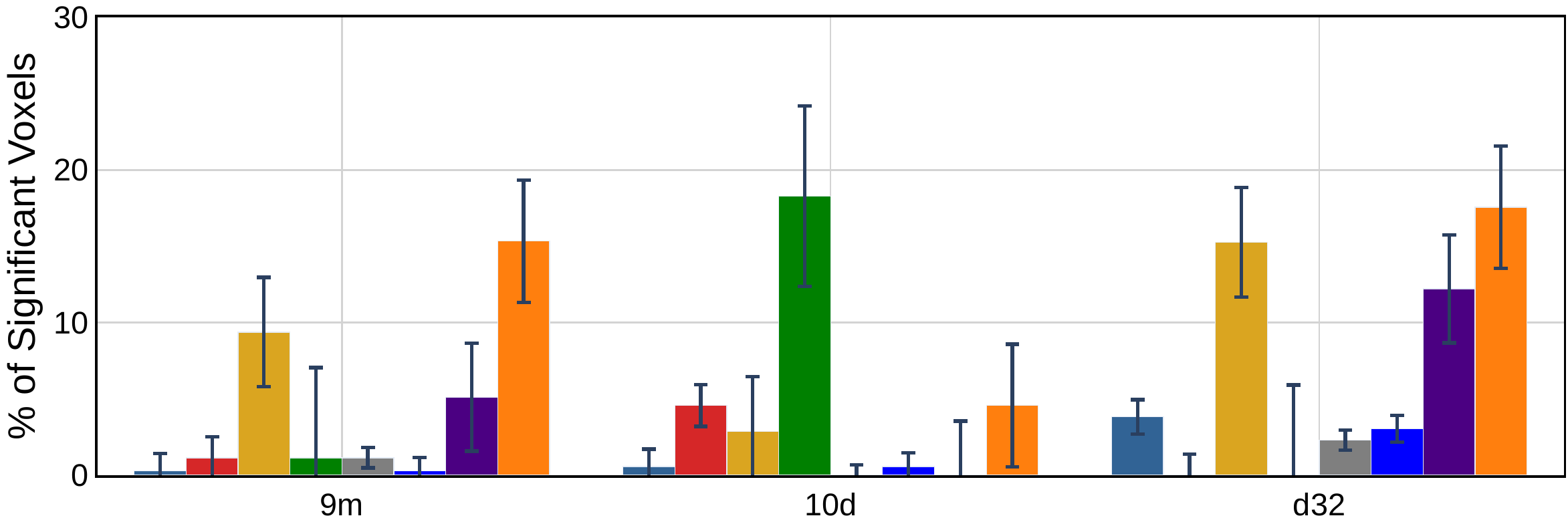}
%\vspace{-1cm}
\caption{\textbf{Performance of Individual Embedding Methods for dmPFC region}: Region of Interest (ROI) analysis of the prediction performance  of various feature sets for various sub-regions of dmPFC region. For each model, we show the \% of ROI voxels with a significant increase in prediction performance. Each bar represents the average score and error bars show standard error across 82 subjects. Left hemisphere (Top) and Right hemisphere (Bottom).}
%‘-’ indicates a hypothesis test for the difference in $R^2$ scores between the two feature groups being larger than 0.}
\label{fig:listening_avg_r2_dmPFC}
\end{figure*}

\begin{figure*}[t] 
\centering
%\vspace{-1cm}
\includegraphics[width=0.85\linewidth]{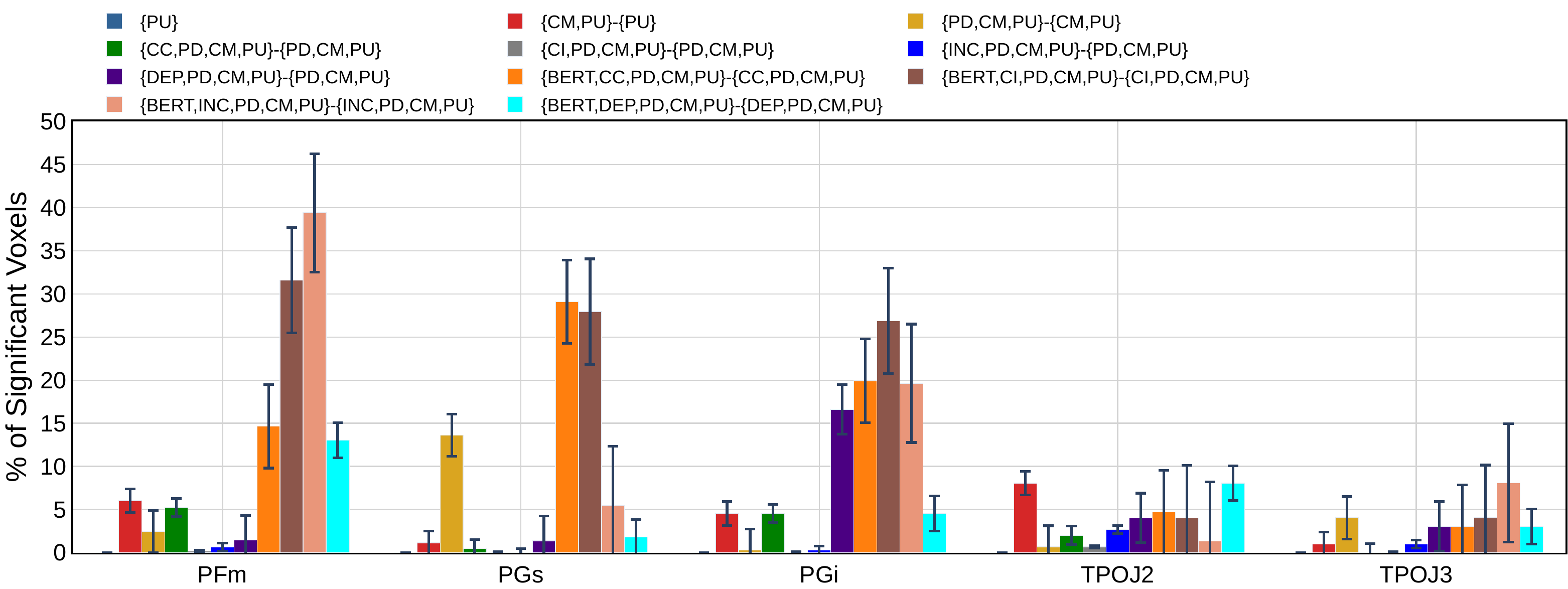}
%\vspace{-0.3cm}
\includegraphics[width=0.85\linewidth]{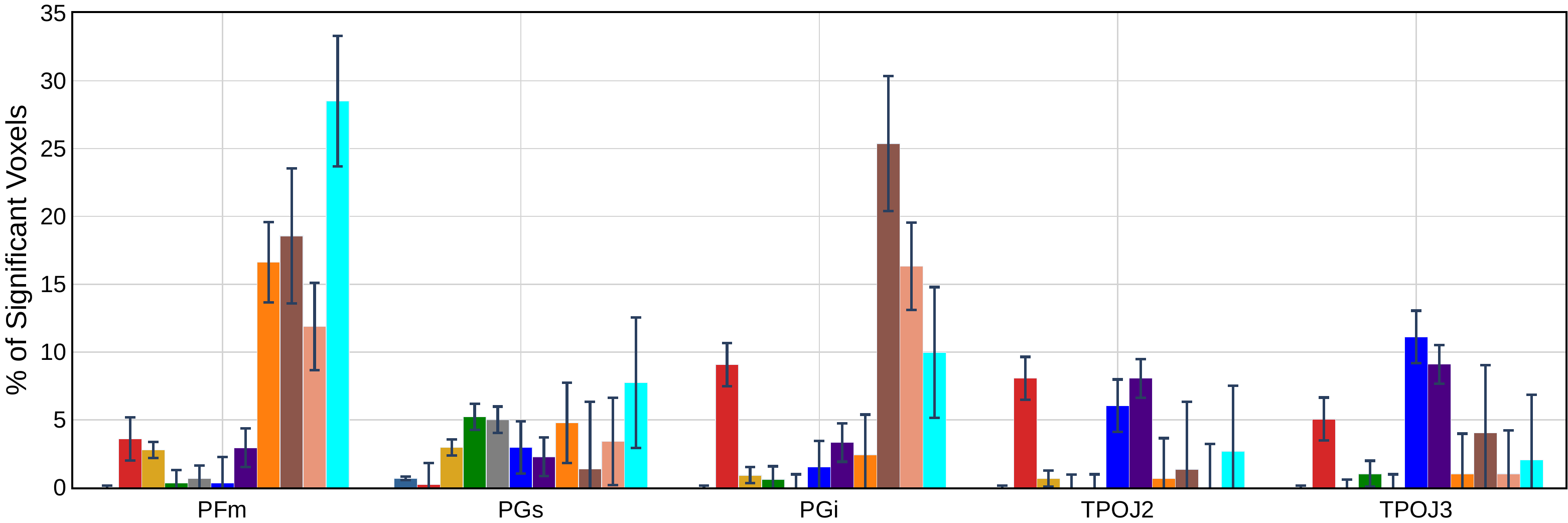}
%\vspace{-1cm}
\caption{\textbf{Additional Predictive Power of various Representations for various sub-regions of Angular Gyrus (AG) region}. For each model, we show the percentage of ROI voxels in which we see a significant increase in prediction performance. Each bar represents the average percentage across 82 subjects, and the error bars show the standard error across subjects. ‘-’ indicates a hypothesis test for the difference in $R^2$ scores between the two feature groups being larger than 0. Left hemisphere (Top) and Right hemisphere (Bottom).}
\label{fig:listening_ag_pairs}
\end{figure*}

\begin{figure*}[t] 
\centering
\vspace{-0.5cm}
\includegraphics[width=0.85\linewidth]{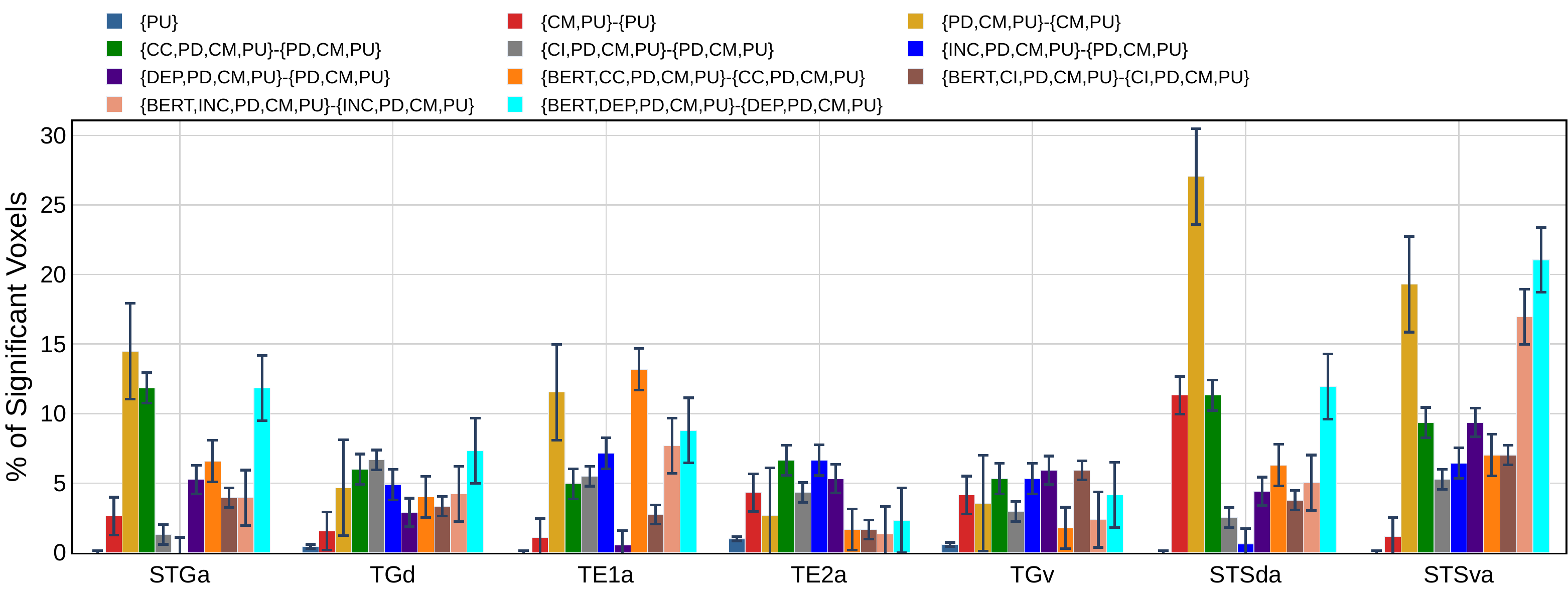}
%\vspace{-0.3cm}
\includegraphics[width=0.85\linewidth]{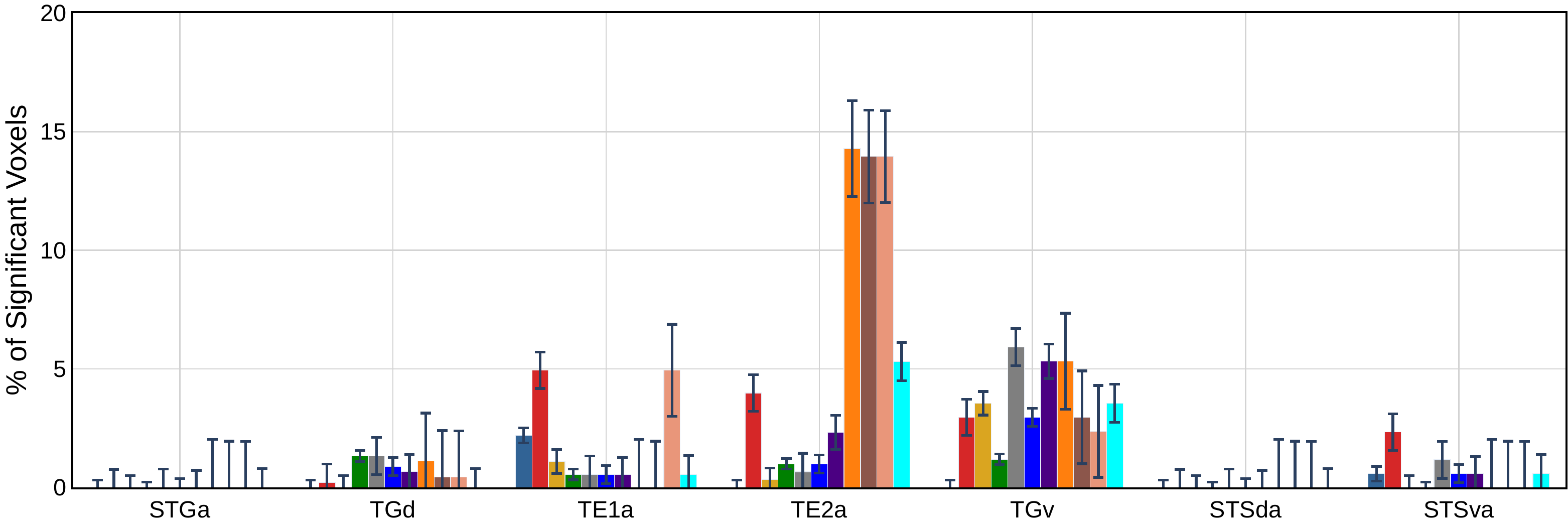}
%\vspace{-1cm}
\caption{\textbf{Additional Predictive Power of various Representations for various sub-regions of Anterior Temporal Lobe (ATL) region}. For each model, we show the percentage of ROI voxels in which we see a significant increase in prediction performance. Each bar represents the average percentage across 82 subjects, and the error bars show the standard error across subjects. ‘-’ indicates a hypothesis test for the difference in $R^2$ scores between the two feature groups being larger than 0. Left hemisphere (Top) and Right hemisphere (Bottom).}
\label{fig:listening_atl_pairs}
\end{figure*}

\begin{figure*}[t] 
\centering
%\vspace{-1cm}
\includegraphics[width=0.85\linewidth]{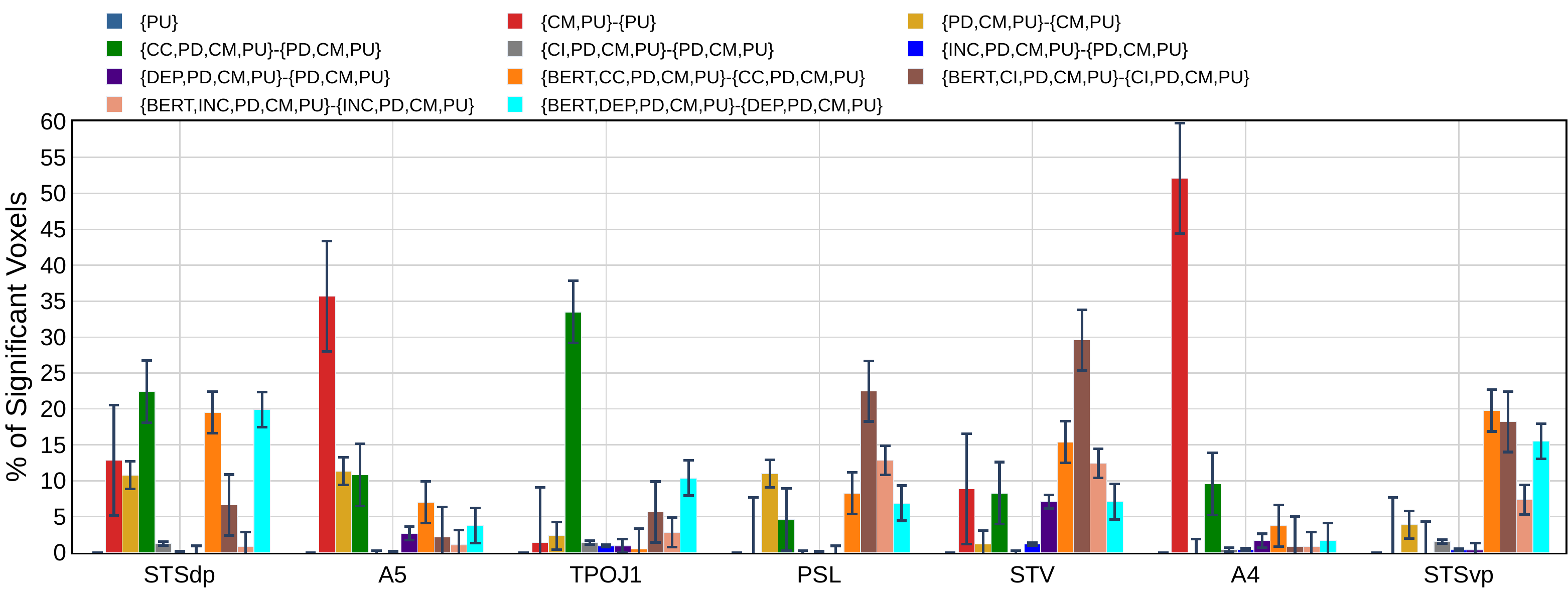}
%\vspace{-0.3cm}
\includegraphics[width=0.85\linewidth]{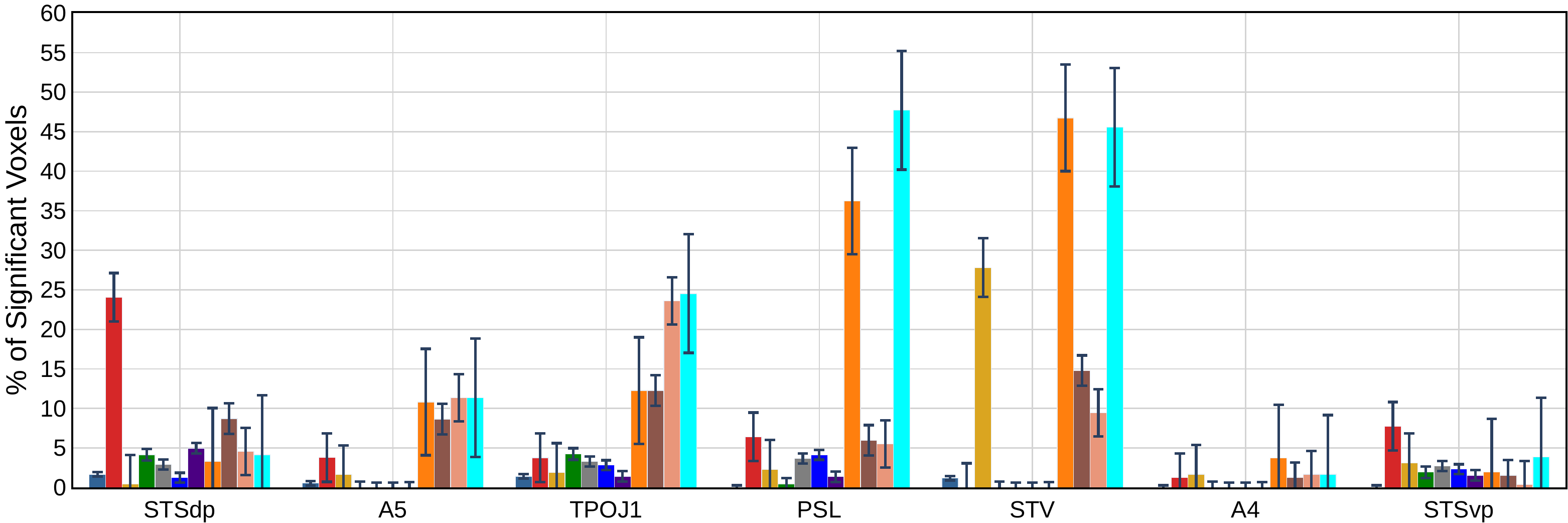}
%\vspace{-1cm}
\caption{\textbf{Additional Predictive Power of various Representations for various sub-regions of Posterior Temporal Lobe (PTL) region}. For each model, we show the percentage of ROI voxels in which we see a significant increase in prediction performance. Each bar represents the average percentage across 82 subjects, and the error bars show the standard error across subjects. ‘-’ indicates a hypothesis test for the difference in $R^2$ scores between the two feature groups being larger than 0. Left hemisphere (Top) and Right hemisphere (Bottom).}
\label{fig:listening_ptl_pairs}
\end{figure*}

\begin{figure*}[t] 
\centering
%\vspace{-1cm}
\includegraphics[width=0.8\linewidth]{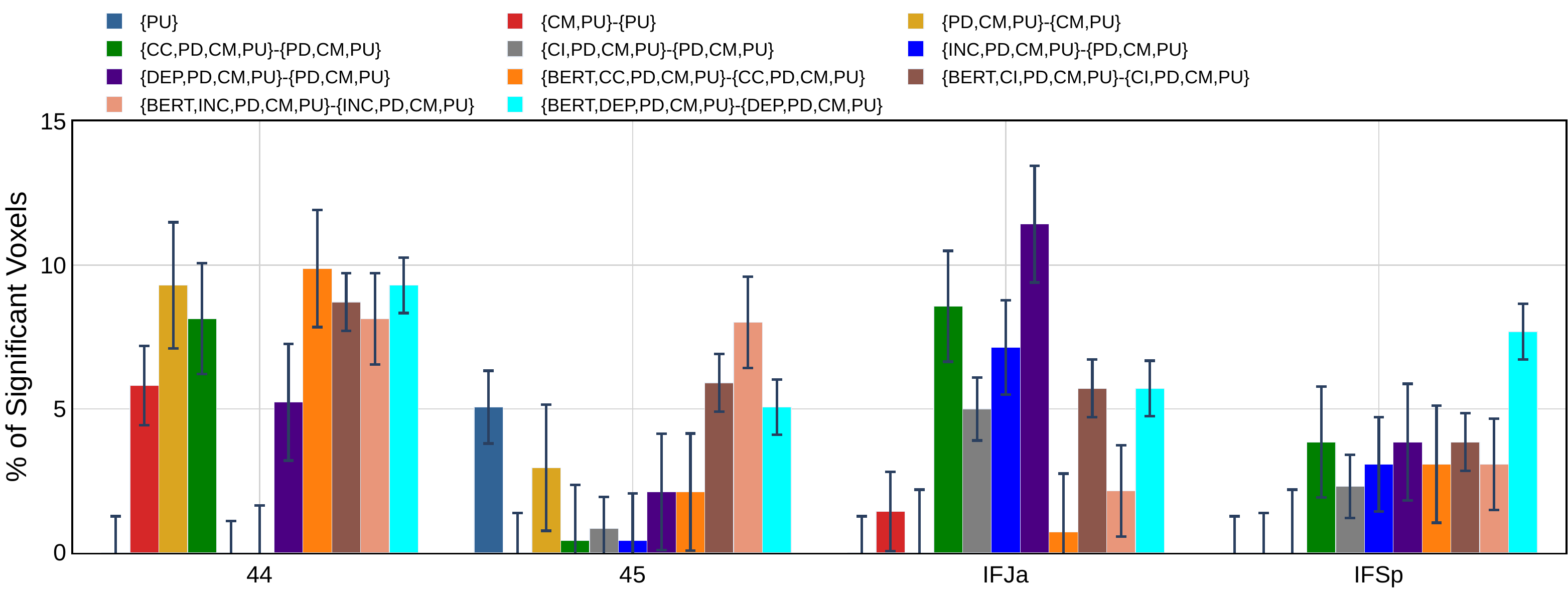}
%\vspace{-0.3cm}
\includegraphics[width=0.8\linewidth]{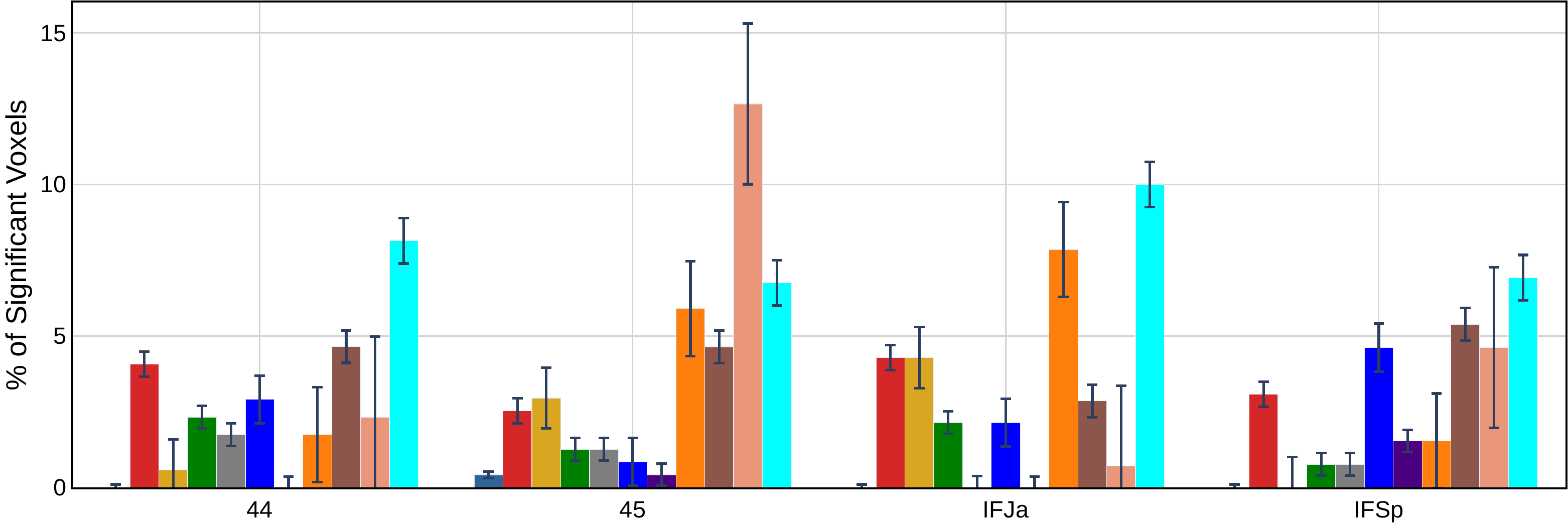}
%\vspace{-1cm}
\caption{\textbf{Additional Predictive Power of various Representations for various sub-regions of Inferior Frontal Gyrus (IFG) region}. For each model, we show the percentage of ROI voxels in which we see a significant increase in prediction performance. Each bar represents the average percentage across 82 subjects, and the error bars show the standard error across subjects. ‘-’ indicates a hypothesis test for the difference in $R^2$ scores between the two feature groups being larger than 0. Left hemisphere (Top) and Right hemisphere (Bottom).}
\label{fig:listening_ifg_pairs}
\end{figure*}

\begin{figure*}[t] 
\centering
%\vspace{-1cm}
\includegraphics[width=0.8\linewidth]{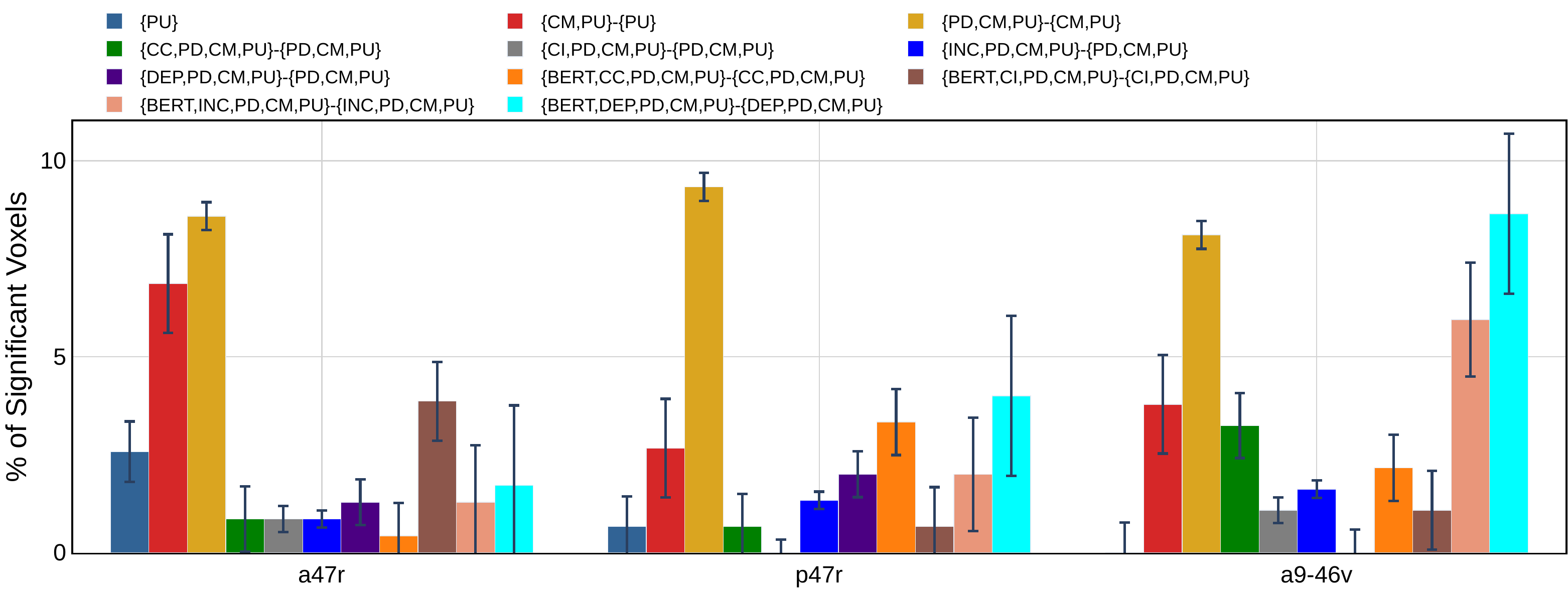}
%\vspace{-0.3cm}
\includegraphics[width=0.8\linewidth]{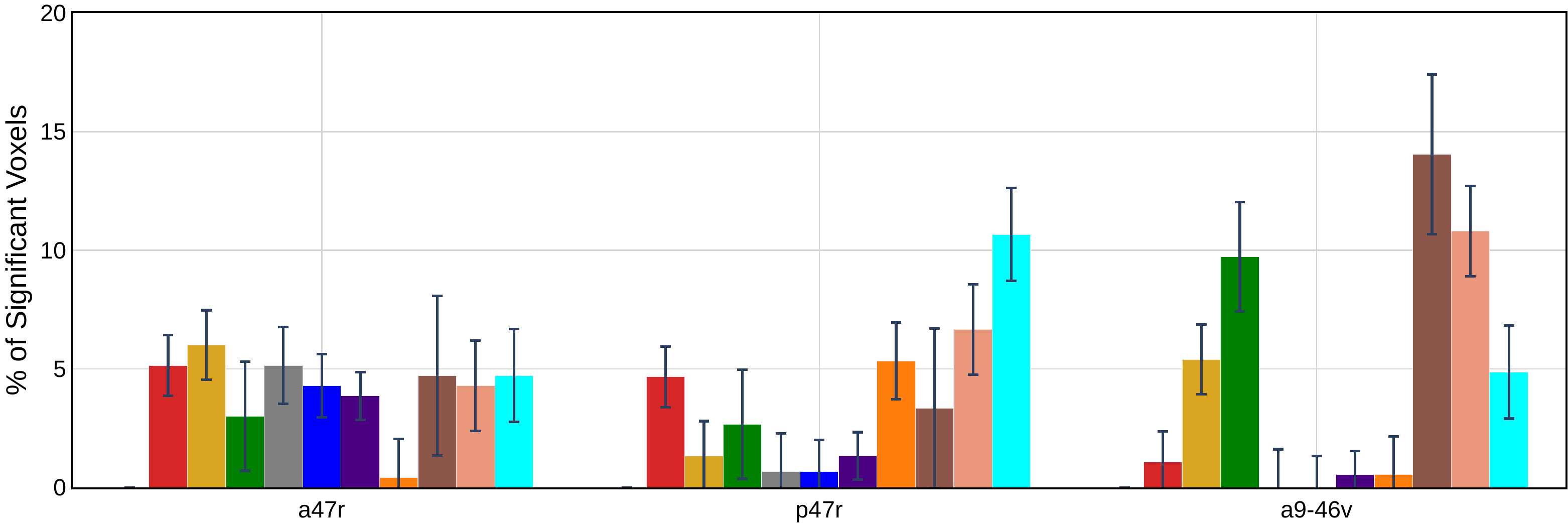}
%\vspace{-1cm}
\caption{\textbf{Additional Predictive Power of various Representations for various sub-regions of Inferior Frontal Gyrus Orbital (IFGOrb) region}. For each model, we show the percentage of ROI voxels in which we see a significant increase in prediction performance. Each bar represents the average percentage across 82 subjects, and the error bars show the standard error across subjects. ‘-’ indicates a hypothesis test for the difference in $R^2$ scores between the two feature groups being larger than 0. Left hemisphere (Top) and Right hemisphere (Bottom).}
\label{fig:listening_ifgorb_pairs}
\end{figure*}

\begin{figure*}[t] 
\centering
%\vspace{-1cm}
\includegraphics[width=0.85\linewidth]{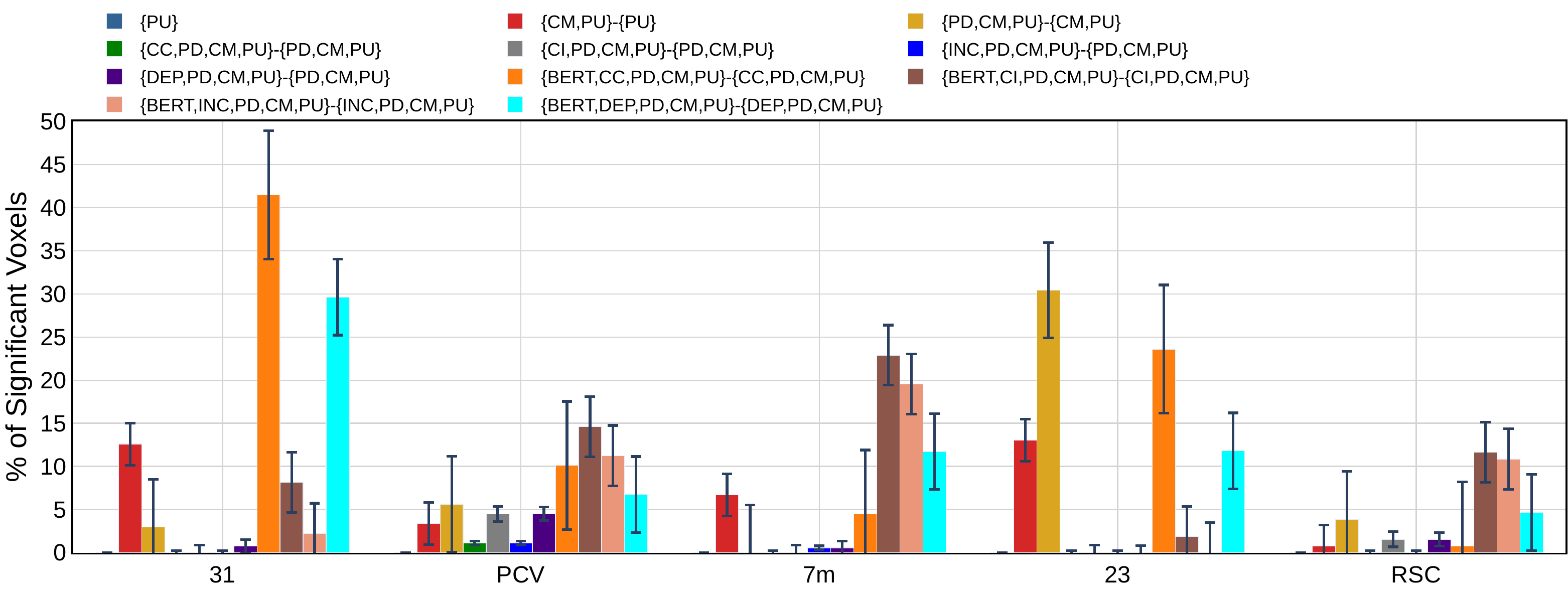}
%\vspace{-0.3cm}
\includegraphics[width=0.85\linewidth]{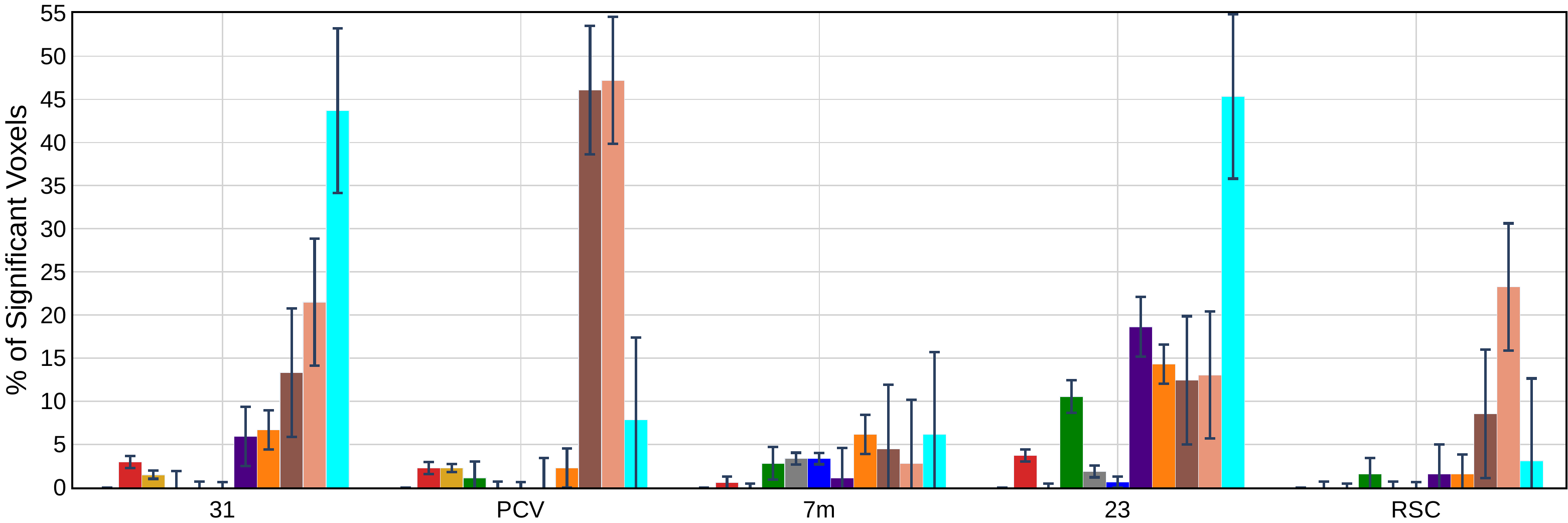}
%\vspace{-1cm}
\caption{\textbf{Additional Predictive Power of various Representations for various sub-regions of Posterior Cingulate Cortex (PCC) region}. For each model, we show the percentage of ROI voxels in which we see a significant increase in prediction performance. Each bar represents the average percentage across 82 subjects, and the error bars show the standard error across subjects. ‘-’ indicates a hypothesis test for the difference in $R^2$ scores between the two feature groups being larger than 0. Left hemisphere (Top) and Right hemisphere (Bottom).}
\label{fig:listening_pcc_pairs}
\end{figure*}

\begin{figure*}[t] 
\centering
%\vspace{-1cm}
\includegraphics[width=0.85\linewidth]{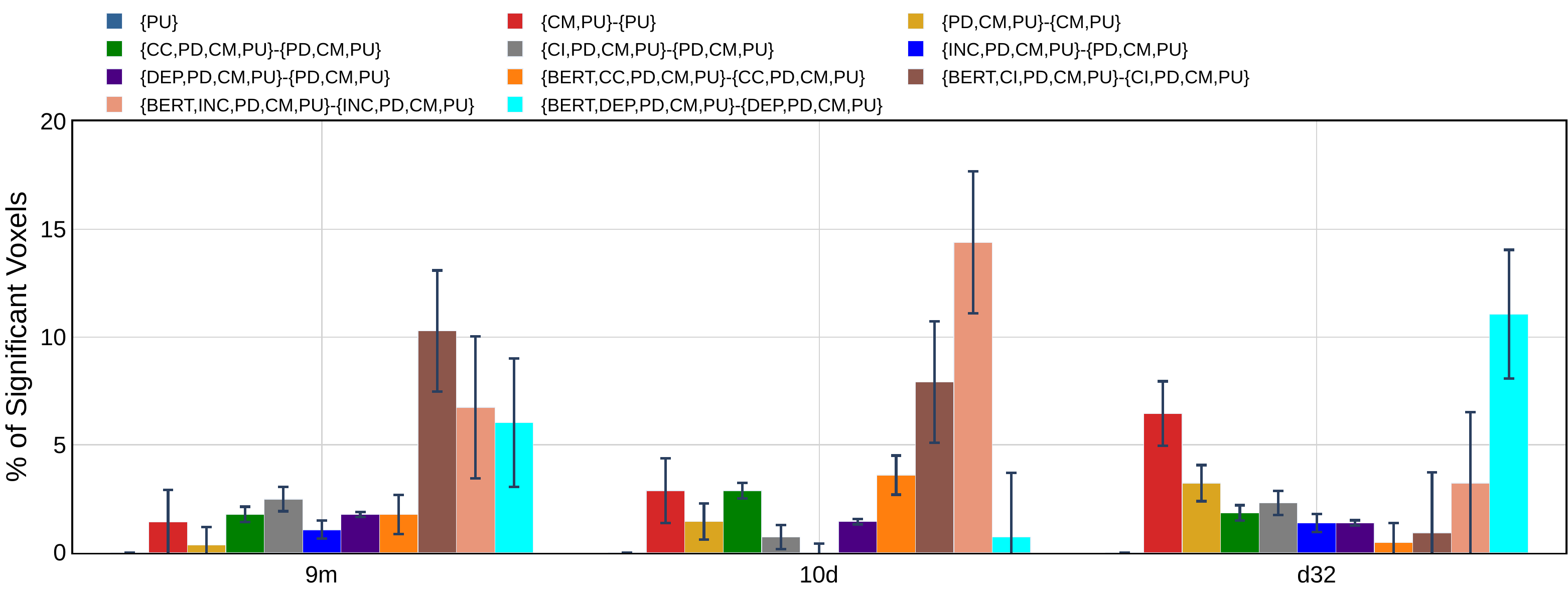}
%\vspace{-0.3cm}
\includegraphics[width=0.85\linewidth]{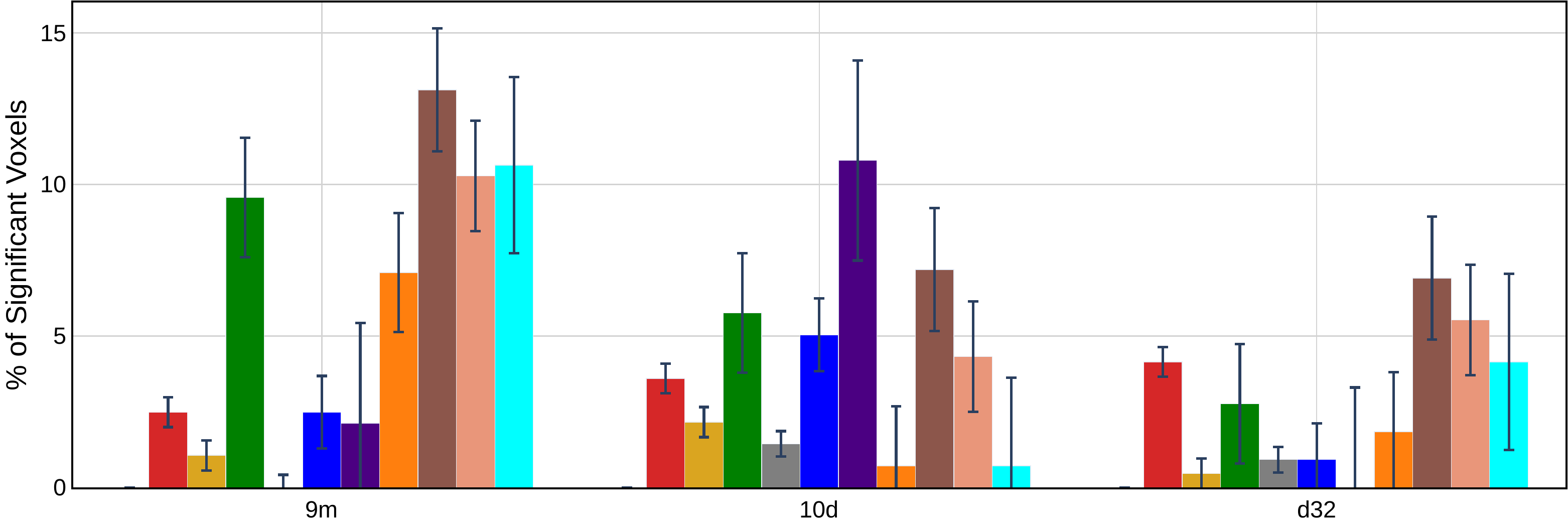}
%\vspace{-1cm}
\caption{\textbf{Additional Predictive Power of various Representations for various sub-regions of Dorsal Medial Prefrontal Cortex (dmPFC) region}. For each model, we show the percentage of ROI voxels in which we see a significant increase in prediction performance. Each bar represents the average percentage across 82 subjects, and the error bars show the standard error across subjects. ‘-’ indicates a hypothesis test for the difference in $R^2$ scores between the two feature groups being larger than 0. Left hemisphere (Top) and Right hemisphere (Bottom).}
\label{fig:listening_dmpfc_pairs}
\end{figure*}

% \begin{figure*}[!htb] 
% \centering
% %\vspace{-1cm}
% \includegraphics[width=0.85\linewidth]{images/narratives_pairs_lefthemisphere.pdf}
% %\vspace{-0.4cm}
% \includegraphics[width=0.85\linewidth]{images/narratives_pairs_righthemisphere.pdf}
% \caption{To be updated (there will be 20 plots here.): Left-Hemisphere (Top) and Right-Hemisphere (Bottom): Region of Interest (ROI) analysis of the prediction performance for pairs of syntactic parsing feature sets and BERT. For each model, we show the percentage of ROI voxels in which we see significant increase in prediction performance. Each bar represents the average percentage across 82 subjects and the error bars show the standard error across subjects. ‘-’ indicates a hypothesis test for the difference in $R^2$ scores between the two feature groups being larger than 0. Note: Groups such as \{CC,SynGCN\}-\{CC\}, \{CI,SynGCN\}-\{CI\}, \{CC,INC\}-\{CC\}, \{CI,INC\}-\{CI\}, \{CC,BERT\}-\{BERT\}, and \{CI, BERT\}-\{CI\} display negligible (<0.5) \% of voxels across all the language regions, and are therefore not plotted.}
% \label{fig:listening_syngcn_bert_r2_left}
% \end{figure*}

% \begin{figure*}[t] 
% \centering
% \includegraphics[width=\linewidth]{images/brainmaps_cc_ci_inc_syngcn.pdf}
% \includegraphics[width=\linewidth]{images/cm_pd_pu_bert.pdf}
% \caption{Pearson correlation between actual and predicted voxels on brain areas for various syntactic feature representations: CC, CI, INC and SynGCN.}
% \label{fig:brainmaps_all}
% \end{figure*}

%\section{Example Appendix}
%\label{sec:appendix}

%This is a section in the appendix.

\end{document}